%% file: main.tex
\documentclass[letterpaper, 10 pt, journal, twoside]{IEEEtran} 

\IEEEoverridecommandlockouts                              

\title{Learning Flexible and Reusable Locomotion Primitives for a Microrobot}
\author{Brian Yang, Grant Wang, Roberto Calandra, Daniel Contreras, Sergey Levine, and Kristofer Pister%
\thanks{Manuscript received: September, 09, 2017; Revised January, 21, 2018; Accepted January, 27, 2018.}
\thanks{This paper was recommended for publication by Editor Yu Sun upon evaluation of the Associate Editor and Reviewers' comments. 
This work was supported by the Berkeley Sensor and Actuator Center, and by Berkeley DeepDrive. \textit{(Brian Yang and Grant Wang contributed equally to this work.) (Corresponding author: \href{mailto:roberto.calandra@berkeley.edu}{Roberto Calandra})}}
\thanks{All the authors are with the Department of Electrical Engineering and Computer Sciences, University of California, Berkeley, USA
        {\tt\small \{brianhyang, grant.wang5, roberto.calandra, dscontreras, ksjp\}@berkeley.edu, svlevine@eecs.berkeley.edu}}%
\thanks{Digital Object Identifier (DOI): 10.1109/LRA.2018.2806083}
}

\input{preamble}
\newcommand{\citep}[1]{\cite{#1}}

\begin{document}

\maketitle


\begin{abstract}
	\input{0_abstract}
\end{abstract}


\section{Introduction}

	\input{1_introduction}


\section{Related Work}
\label{sec:related}

	\input{2_related}


\section{The Hexapod Microrobot}
	
    \input{3_hardware}


\section{Background}
\label{sec:background}

	\input{4_background}


\section{Learning Locomotion Primitives for Path Planning}
\label{sec:approach}

\input{5_approach}


\section{Experimental Simulation Results}
\label{sec:results}

\input{6_results}


\section{Conclusion}

	\input{7_conclusion}


%


\bibliographystyle{IEEEtran}
\bibliography{paper-nanorobots}  

\end{document}

%% file: preamble.tex
\usepackage[english]{babel}

\usepackage{graphicx}
\usepackage{animate}
\usepackage{epsfig} 				
\usepackage{epstopdf}

\usepackage{listings}
\usepackage{color}
\usepackage{nameref}
\usepackage{hyperref}
\usepackage[colorinlistoftodos]{todonotes}
\usepackage{amsmath}	 			
\usepackage{amssymb}  				

\usepackage{dsfont}			
\usepackage{mathtools}

\usepackage{epigraph}
\usepackage{lscape}
\usepackage[]{nomencl}				
\usepackage{algorithm}
\usepackage{algorithmic}
\usepackage{multicol}
\usepackage{multirow}
\usepackage{etoolbox}

\usepackage{caption}
\usepackage{subcaption}

\usepackage{wrapfig}
\usepackage{multimedia}

\usepackage{siunitx}
\usepackage{flushend}

\usepackage{floatflt}

\usepackage{url}

\usepackage[absolute,overlay]{textpos}

\usepackage{bibentry}

\usepackage{tikz}
\usepgflibrary{arrows}	

\usepackage{float}
\usepackage[utf8]{inputenc}
\usepackage[english]{babel}
\usepackage{graphics}
\usepackage{animate}

\usepackage{listings}

\usepackage[nostamp]{draftwatermark}

\usepackage{extarrows}		

\allowdisplaybreaks[1]

\newcommand{\playvideo}[1]{\href{run:#1}{\includegraphics[scale=0.12]{\RCPath fig/empty}}}

\usepackage{lipsum}
\usepackage{framed}

\input{commands.tex}

\AtEndDocument{\par\leavevmode}

%% file: commands.tex
\newcommand{\ie}[0]{i.e.}
\newcommand{\eg}[0]{e.g.}

\newcommand{\email}[1]{\href{mailto:#1}{\nolinkurl{#1}}}

\newcommand{\link}[1]{\colora{\url{#1}}}

\renewcommand{\sec}[1]{Section~\ref{#1}}
\newcommand{\fig}[1]{Figure~\ref{#1}}
\newcommand{\eq}[1]{Equation~\eqref{#1}}

\definecolor{matlab1}{rgb}{0,0,1}
\definecolor{matlab2}{rgb}{0,0.5,0}
\definecolor{matlab3}{rgb}{1,0,0}
\definecolor{matlab4}{rgb}{0,0.75,0.75}
\definecolor{matlab5}{rgb}{0.75,0,0.75}
\definecolor{matlab6}{rgb}{0.75,0.75,0}
\definecolor{matlab7}{rgb}{0.25,0.25,0.25}

\definecolor{darkgreen}{rgb}{0,0.5,0}		
\definecolor{purple}{rgb}{0.75,0,0.75}
\definecolor{pink}{rgb}{1,0.4,0.6}

\newcommand{\capitalize}[1]{\expandafter\MakeUppercase\expandafter{#1}}

\newcommand{\colora}[1]{{\usebeamercolor[fg]{framesubtitle}#1}}

\makeatletter
\newcommand*{\compress}{\@minipagetrue}
\makeatother

\renewcommand{\vec}[1]{\boldsymbol{#1}}				

\newcommand{\dataset}[0]{\mathcal{D}}

\newcommand{\maximize}[0]{\text{arg max}} 			

\newcommand{\parameter}[0]{\theta} 				
\newcommand{\parameters}[0]{\vec{\parameter}} 			
\newcommand{\context}[0]{\vec{c}}
\newcommand{\objfuncNo}[0]{f}					
\newcommand{\objfunc}[1]{\objfuncNo\left(#1\right)}		

\newcommand{\numbersubobj}[0]{n}

\newcommand{\dom}[0]{\succ} 

\newcommand{\q}{\vec{q}}					
\ifdef{\dq}{\renewcommand{\dq}{\dot{\q}}}{\newcommand{\dq}{\dot{\q}}}

\usepackage{array}
\makeatletter
\newcommand{\thickhline}{%
    \noalign {\ifnum 0=`}\fi \hrule height 1pt
    \futurelet \reserved@a \@xhline
}
\newcolumntype{"}{@{\hskip\tabcolsep\vrule width 1pt\hskip\tabcolsep}}
\makeatother

%% file: 0_abstract.tex
The design of gaits for robot locomotion can be a daunting process which requires significant expert knowledge and engineering.
This process is even more challenging for robots that do not have an accurate physical model, such as compliant or micro-scale robots.
Data-driven gait optimization provides an automated alternative to analytical gait design.
In this paper, we propose a novel approach to efficiently learn a wide range of locomotion tasks with walking robots.
This approach formalizes locomotion as a contextual policy search task to collect data, and subsequently uses that data to learn multi-objective locomotion primitives that can be used for planning.
As a proof-of-concept we consider a simulated hexapod modeled after a recently developed microrobot, and we thoroughly evaluate the performance of this microrobot on different tasks and gaits.
Our results validate the proposed controller and learning scheme on single and multi-objective locomotion tasks.
Moreover, the experimental simulations show that without any prior knowledge about the robot used (e.g., dynamics model), our approach is capable of learning locomotion primitives within 250 trials and subsequently using them to successfully navigate through a maze.

%% file: 1_introduction.tex
\IEEEPARstart{S}{ubstantial} progress has been made in recent years towards the development of fully autonomous microrobots~\citep{saito_miniaturized_2016,Vogtmann2017}.
However, gait design for robot locomotion at the sub-centimeter scale is not a well-studied problem.
Completing more complicated locomotion tasks like navigating complex environments is even more challenging.
These issues become exacerbated when dealing with legged locomotion, where even walking straight is still an active area of study for normal-sized robots.
In this paper, we present a novel approach for the autonomous optimization of locomotion primitives and gaits.

While locomotion on larger-scale robots has been thoroughly investigated, transferring many of these proven approaches to the millimeter scale poses many unique challenges.
One such obstacle is the lack of access to sufficiently accurate simulated models at the millimeter scale.
Even simulation environments designed to simulate dynamics at this scale are generally unequipped for usage in robotics contexts.
Additionally, working with microrobots can place severe limitations on the number of iterations as trials become much more time-consuming and expensive to run.

While microrobot locomotion has been addressed in the past, much of the work is primarily concerned with the mechanical design and manufacturing of microrobots.
Accomplishing more sophisticated locomotion tasks on the sub-centimeter scale remains an open area for research.
Analytical implementations of various gait behaviors have worked on microrobots~\citep{ebefors_walking_1999,hollar_solar_2003}, but these solutions can become unwieldy for robots with higher DOF such as legged walkers (\eg, our micro-hexapod).
Data-driven automatic gait optimization is a viable alternative to analytical gait design and optimization, but using these techniques can be challenging due to the high number of trials that might be necessary to perform in order to learn viable gaits.

\begin{figure}[t]
  \centering
  \includegraphics[width=0.94\linewidth]{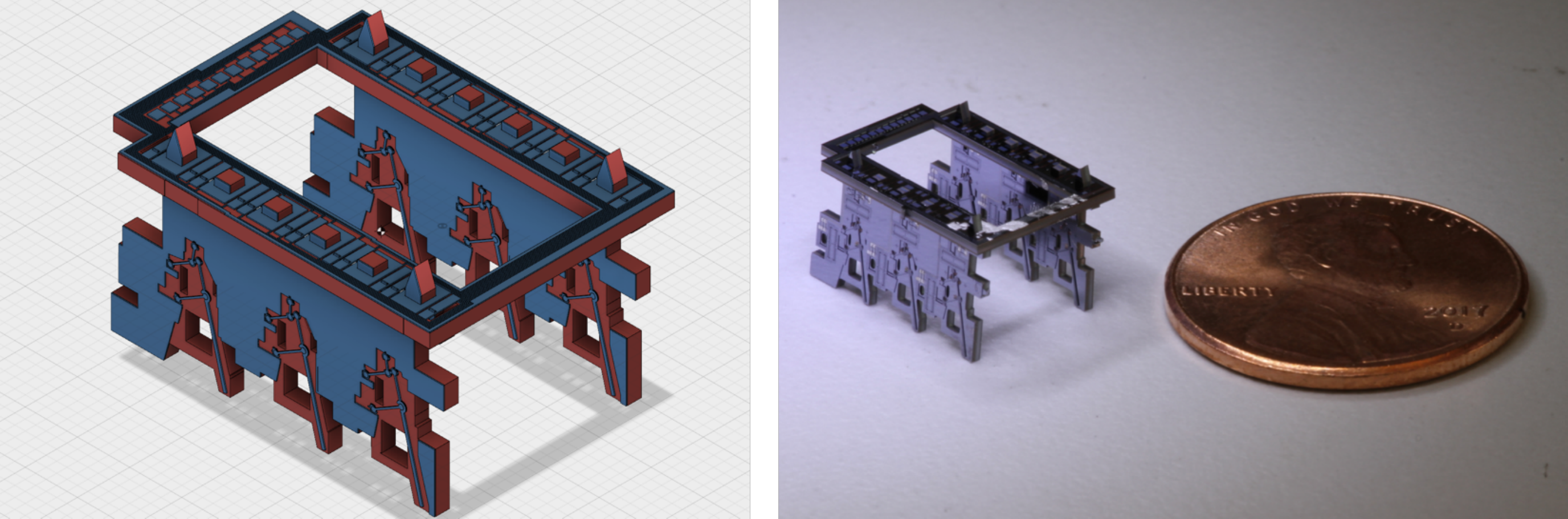}
  \caption{The six-legged micro walker considered in our study as a CAD model (left) and an assembled prototype (right).}
  \label{fig:flik_cad}
 \vspace{-10pt}
\end{figure}

Our contributions are two-fold:
1) we validate the use of both CPG controllers and Bayesian optimization for microrobots on a wide range of single and multi-objective locomotion tasks.
2) we introduce a novel approach to efficiently learn gaits and motor primitives from scratch without the need for prior knowledge (\eg, a dynamics model).
This is accomplished by collecting data on various motor primitives using contextual policy search and using those evaluations to reformulate the problem into a multi-objective optimization task, providing us a model that can map any set of parameters to a predicted trajectory.
Using this model, we can optimize our parameters on various trajectories for subsequent use in path planning.
This approach is not tied exclusively to microrobots, but can be used for any walking robot.

To evaluate our approach, we used a simulated hexapod microrobot modeled after the recently developed microrobot~\citep{contreras_first_2017} shown in \fig{fig:flik_cad}. 
We first validate the use of a CPG controller on our microrobot to reduce the number of parameters during optimization.
Then, we validate the use of Bayesian optimization and existing techniques on a curriculum of progressively more difficult tasks including learning single-objective, contextual, and multi-objective gaits.
As a proof of concept, we evaluated our approach by learning motor primitives from 250 trials and subsequently using them to successfully navigate through a maze.

%% file: 2_related.tex
There has been an abundance of work published on the design and development of walking~\citep{Ambroggi1997} and flying millimeter-scale microrobots~\citep{4457871,8001934, Kilberg2017MEMSAC}.
Much of this work focuses on hardware considerations such as the design of micro-sized joints and actuators rather than control.
To our knowledge, no previous work has implemented a CPG-based controller for on-board control of a walking microrobot, nor has learning been used for locomotion on a microrobots.

While hexapod gaits have been thoroughly studied and tested~\citep{Altendorfer2001,Hoover2008}, much of the work did not easily transfer to our microrobot due to the drastically different leg dynamics. 
Most hexapods make use of rotational joints with higher DOF while our walker uses only two prismatic spring joints per leg, resulting in less control and unique constraints on leg retraction and actuation. 

While sufficient for simple controllers with few parameters, manually tuning controller parameters can require an immense amount of domain expertise and time.
As such, automatic gait optimization is an important research field that has been studied with a wide variety of approaches in both the single-objective~\citep{Tedrake2004,Chernova2004,Niehaus2007,Lizotte2007,Tesch2011,Oliveira2011,Oliveira2013,Calandra2015a} and multi-objective setting~\citep{Capi2005,Oliveira2011,Oliveira2013,Tesch2013}.
Evolutionary algorithms have been successfully used to train quadrupedal robots~\citep{Chernova2004,Oliveira2011}, but this approach often requires thousands of experiments before producing good results, which is unfeasible on fragile microrobots.

A more data-efficient approach used before to learn gaits for snake and bipedal robots is Bayesian optimization~\citep{Lizotte2007,Tesch2011,Calandra2015a,Antonova2017}. 
Bayesian optimization has been applied to contextual policy search in the context of robot manipulation~\citep{Metzen2015}.
Our contribution builds off of this work by applying and extending the contextual framework to learning movement trajectories and path planning.
%
Another extension of Bayesian optimization related to our work is Multi-objective Bayesian optimization, which has also been previously applied in the context of robotic locomotion~\cite{Tesch2013}.
However, past work is only concerned with using multi-objective optimization to balance the trade-off between various competing goals.
Our main contribution demonstrates an entirely novel application of multi-objective optimization to learning motor primitives that does not involve the trade-off between various goals, but instead uses a multi-objective model to learn over an area of possible trajectories for path planning.

%% file: 3_hardware.tex
We now introduce the hexapod microrobot considered in this paper.
This robot is of particular interest due to the unique challenges that arise when attempting traditional gait design techniques.
The micro-scale of the walker makes it very challenging to obtain an accurate dynamics model.
Moreover, the robot is subject to wear-and-tear, and therefore any learning approach employed must be capable of learning gaits within a limited number of trials.

\subsection{Physical Description}    
	\begin{figure}[t]
	  \centering
	  \includegraphics[width=0.96\linewidth]{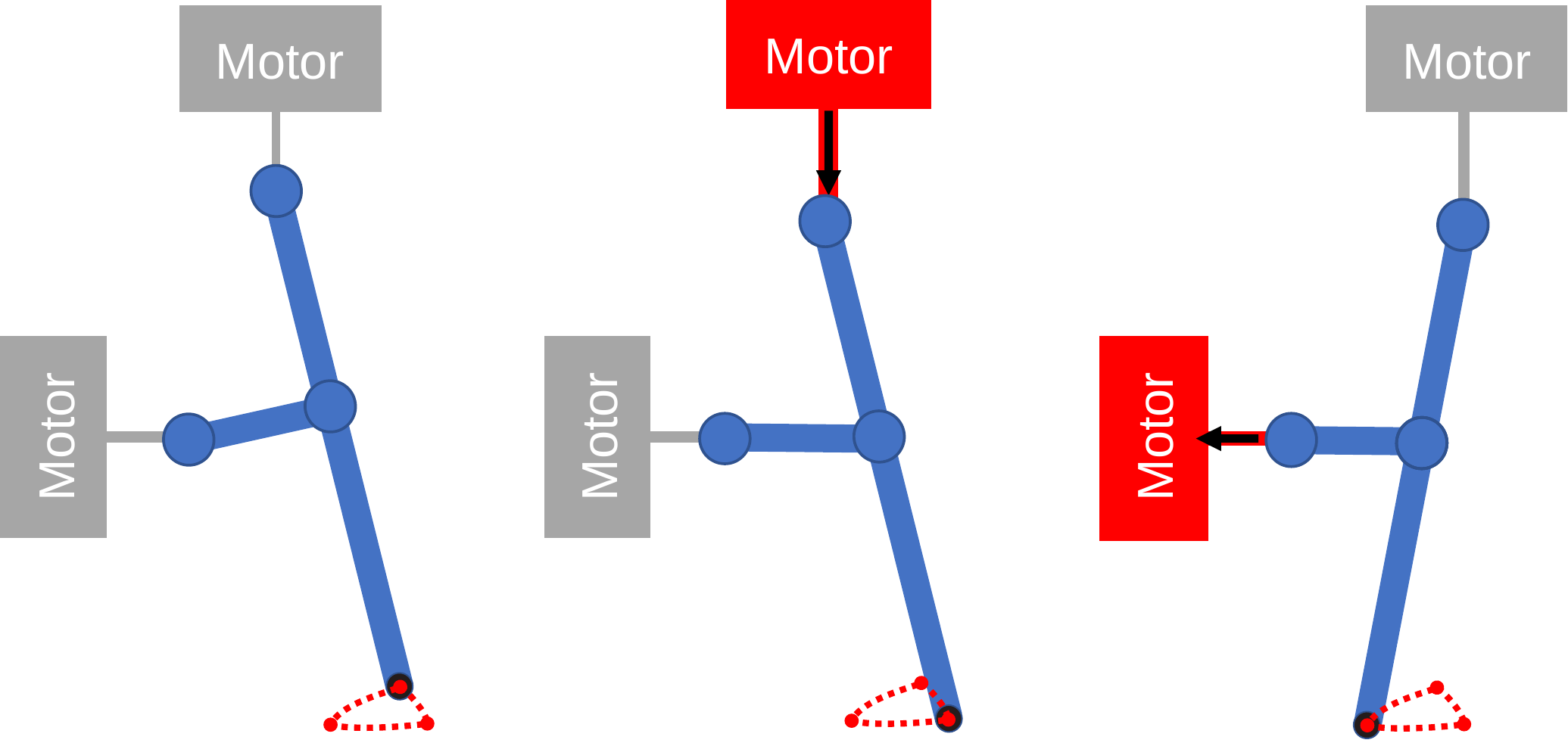}
	  \caption{Diagram of the robot leg showing the actuation sequence (active motors are shown in red). Each leg has 2 motors, each one independently actuating a single DOF. 
      }
	  \label{fig:leg_diagram}
	\end{figure}

	The hexapod microrobot is based on silicon microelectromechanical systems (MEMS) technology. 
	The robot's legs are made using linear motors actuating planar pin-joint linkages~\citep{Contreras2016DurabilityOS}. 
	A tethered single-legged walking robot was previously demonstrated using this technology~\citep{contreras_first_2017}. 
	The hexapodal robot is assembled using three chips.  
	The two chips on the side each have 3 of the leg assemblies, granting six 2 degree-of-freedom (DOF) legs for the whole robot. 
	The top chip acts to hold the leg chips together for support, and to route the signals for off-board power and control.
	Overall, the robot measures \SI{13}{\milli\meter} long by \SI{9.6}{\milli\meter} wide and stands at \SI{8}{\milli\meter} tall with an overall weight of approximately \SI{200}{\milli\gram}. 
 
\subsection{Actuation}
	Each of the robot's legs has 2-DOF in the plane of fabrication, as shown in \fig{fig:leg_diagram}.
	Both DOFs are actuated, thus the leg has 2 motors, one to actuate the vertical DOF to lift the robot's body and a second to actuate the horizontal DOF for the vertical stride. 
	The actuators used for the legs are electrostatic gap-closing inchworm motors~\cite{penskiy_optimized_2013}.
	During a full cycle, each leg moves \SI{0.6}{\milli\meter} vertically with a horizontal stride of \SI{2}{\milli\meter}. 
	For more details on the actuation mechanism used on our microrobot, we refer readers to ~\cite{Contreras2017DynamicsOE}.
	
\subsection{Simulator}
	\begin{wrapfigure}{r}{0.46\linewidth} 
	  \centering
	  \vspace{-10pt}
	  \includegraphics[width=0.98\linewidth]{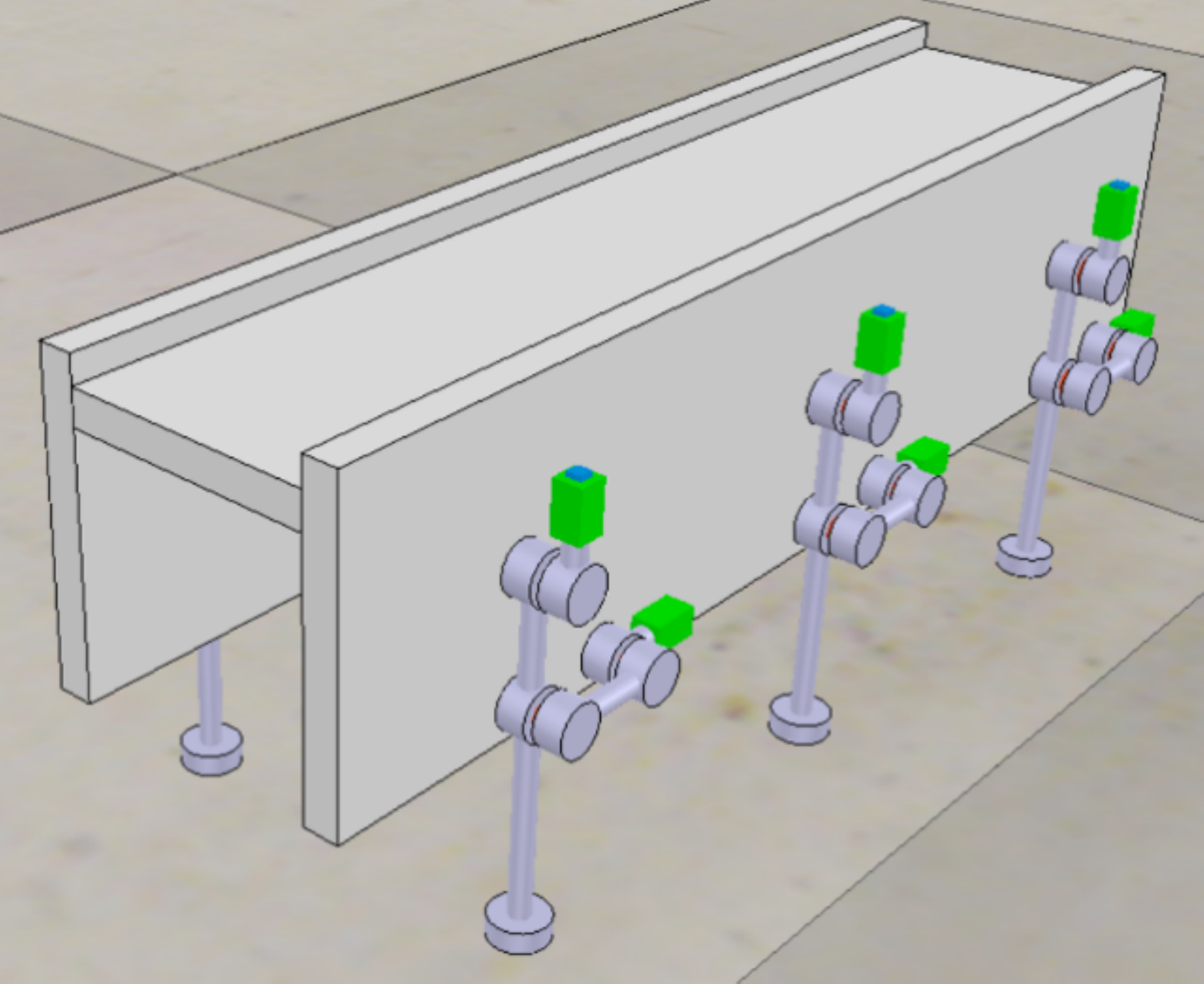}
	  \caption{The simulated micro walker.}
	  \label{fig:vrep}
	  \vspace{-8pt}
	\end{wrapfigure}
	In our experimental simulations, we used the robotics simulator V-REP~\citep{vrep} for constructing a scaled-up simulated model of the physical microrobot (see \fig{fig:vrep}).
	Since V-REP was not designed with simulation of microrobots in mind, it was not capable of simulating the dynamics of the leg joints accurately and would produce wildly unstable models at the desired scale.
	We chose to scale up the size of the robot in simulation by a factor of $100$ in order to account for the issues with scaling in simulation (all the experimental results are re-normalized to the dimensions of the real robot).
	We believe that this re-scaling still allows meaningful results to be produced for several reasons.
	First, the experiments performed in this paper are meant to demonstrate the validity of the proposed controller, and the learning approach for training an actual physical microrobot.
	The policies trained are not meant to work on the real robot without any re-tuning or modification.
	Second, the simulator still allows to test the basic motion patterns we want to implement on the microrobot.
	Finally, our contribution lends credibility to the potential application of Bayesian-inspired optimization methods to a setting where evaluations can be costly and time consuming.

%% file: 4_background.tex
\subsection{Central Pattern Generators}
\label{sec:bg:cpg}
	Central pattern generators (CPGs) are neural circuits found in nearly all vertebrates, which produce periodic outputs without sensory input~\citep{Junzhi_Yu_2014}.
	CPGs are also a common choice for designing gaits for robot locomotion~\citep{Ijspeert2008}.
    We chose to use CPGs for our controller because they are capable of reproducing a wide variety of different gaits simply by manipulating the relative coupling phase biases between oscillators. 
	This allows us to easily produce a variety of gait patterns without having to manually program those behaviors. 
	In addition, CPGs are not computationally intensive and can have on-chip hardware implementations using VLSI or FPGA. 
	This makes them well suited to be eventually used in our physical microrobot, where the processing power is limited.
	CPGs can be modeled as a network of coupled non-linear oscillators where the dynamics of the network are determined by the set of differential equations
	\begin{align}
	  \dot{\phi_i} &= \omega_{i} + \displaystyle \sum_{j} (\omega_{ij}r_j\sin(\phi_j - \phi_i - \varphi_{ij}))\,,\\
	  \ddot{r_i} &= a_r (\dfrac{a_r}{4} (R_i - r_i) - \dot{r_i})\,,\\
	  \ddot{x_i} &= a_x (\dfrac{a_x}{4} (X_i - x_i) - \dot{x_i})\,,
	\end{align}
	where $\phi_{i}$ is a state variable corresponding to the phase of the oscillations and $\omega_{i}$ is the target frequency for the oscillations. 
    $\omega_{ij}$ and $\varphi_{ij}$ are the coupling weights and phase biases which change how the oscillators influence each other.
    To implement our desired gaits, we only need to modify the phase biases between the oscillators~$\phi_{ij}$.
    $r_{i}$ and $x_{i}$ are state variables for the amplitude and offset of each oscillator, and $R_{i}$ and $X_{i}$ are control parameters for the desired amplitude and offset. 
    The constants $a_{r}$ and $a_{x}$ are constant positive gains and allow us to control how quickly the amplitude and offset variables can be modulated. 
    A more detailed explanation of the network can be found in Crespi's original work~\citep{Crespi_2007}. 
	One of the foremost benefits of using a CPG controller is a drastic reduction in the number of parameters~$\parameters_i$ we need to optimize.
    Overall, the parameters that we consider during the optimization are $\parameters = \left[\omega, R, X_{l}, X_{r} \right]$ where $\omega$ is the frequency of the oscillators and $R$ is the phase difference between each of the vertical-horizontal oscillator pairs. 
    In order to allow for directional control, $X_{l}$ and $X_{r}$ are the amplitudes of the left and right side oscillators respectively.    
    
\subsection{Bayesian Optimization}
\label{sec:bg:BO}
	Even with a complete CPG network, some amount of parameter tuning is necessary to obtain efficient locomotion. 
	To automate the parameter tuning, we use Bayesian optimization (BO), an approach often used for global optimization of black box functions~\citep{Kushner1964,Jones2001,Calandra2015a}.
	We formulate the tuning of the CPG parameters as the optimization
	\begin{align}
		\parameters^* = \maximize_{\parameters}\, \objfunc{\parameters}\,,
		\label{eq:optimization}
	\end{align}
	where $\parameters$ are the CPG parameters to be optimized w.r.t. the objective function of choice~$\objfuncNo$ (\eg, walking speed, which we investigate in \sec{sec:results:soo}).
	At each iteration, BO learns a model $\tilde{\objfuncNo}: \parameters \rightarrow \objfunc{\parameters}$ from the dataset of the previously evaluated parameters and corresponding objective values measured~$\dataset=\{\parameters, \objfunc{\parameters}\}$.
	Subsequently, the learned model $\tilde{\objfuncNo}$ is used to perform a ``virtual'' optimization through the use of an acquisition function which controls the trade-off between exploration and exploitation.
	Once the model is optimized, the resulting set of parameters $\parameters^*$ is finally evaluated on the real system, and is added to the dataset together with the corresponding measurement $\objfunc{\parameters^*}$  before starting a new iteration.
	A common model used in BO for learning the underlying objective, and the one that we consider, is Gaussian processes~\citep{Rasmussen2006}. 
	For more information regarding BO, we refer the readers to~\citep{Jones2001,Shahriari2016}.

\subsection{Multi-objective Bayesian Optimization}
	A special case of the optimization task of \eq{eq:optimization} is multi-objective optimization~\citep{Branke2008a}.
	Often times in robotics\footnote{As well as in nature~\citep{Hoyt1981}.}, there are multiple conflicting objectives that need to be optimized simultaneously, resulting in design trade-offs (e.g., walking speed vs energy efficiency which we investigate in \sec{sec:results:moo}).  
	When multiple objectives are taken into consideration, there is no longer necessarily a single optimum solution, but rather the goal of the optimization became to find the set of Pareto optimal solutions~\citep{Pareto1906}, which also takes the name of Pareto front~(PF).
	Formally, the PF is the set of parameters that are not dominated, where a set of parameters~$\parameters_1$ is said to dominate $\parameters_2$ when 
	\begin{align}
		\left\{
		\begin{array}{l l}
		\forall i \in \{1,\dots ,\numbersubobj\}: &\objfuncNo_i(\parameters_1) \leq \objfuncNo_i(\parameters_2)\\
		\exists j \in \{1,\dots , \numbersubobj\}:  &\objfuncNo_j(\parameters_1) < \objfuncNo_j(\parameters_2)
		\end{array} \right.
	\end{align}
	Intuitively, if $\parameters_1 \dom \parameters_2$, then $\parameters_1$ is preferable to $\parameters_2$ as it never performs worse, but at least in one objective function it performs strictly better. 
	However, different dominant variables are equivalent in terms of optimality as they represent different trade-offs.
	      
	Multi-objective optimization can often be difficult to perform as it might require a significant amount of experiments.
	This is especially true with our microrobot where large number of experiments can wear-and-tear the robot.
	As a result, the number of evaluations allowed to find the Pareto set of solutions is limited. 
	Luckily for us, there exist extensions of BO which address multi-objective optimization.
	In particular, the multi-objective Bayesian optimization algorithm that we consider is ParEGO~\citep{Knowles2006}. 
	The main intuition of ParEGO is that at every iteration, the multiple objectives can be randomly scalarized into a single objective (via the augmented Tchebycheff function), which is subsequently optimized as in the standard Bayesian optimization algorithm (by creating a response surface, and then optimizing its acquisition function).
	For more information about multi-objective Bayesian optimization we refer the reader to~\citep{Wagner2010}.
    
\subsection{Contextual Bayesian Optimization}
	Another special case of the optimization task of \eq{eq:optimization}, is contextual optimization.
	In contextual optimization, we assume that there are multiple correlated, but slightly different, tasks which we want to solve, and that they are identified by a context variable~$\context$.
	An example (which we investigate in \sec{sec:results:context1}) might be walking on inclined slopes, where the contextual variable is the angle of the slope.
	The contextual optimization can hence be formalized as 
	\begin{align}
		\parameters^* = \maximize_{\parameters}\, \objfunc{\parameters,\context}\,,
	\end{align}
	where for each context~$\context$, a potentially different set of parameters~$\parameters^*$ exists.
	The main advantage compared to treating each task independently is that, in contextual optimization, we can exploit the correlation between the tasks to generalize, and as a result quickly learn how to solve a new context.
	Specifically, in this paper we consider contextual Bayesian optimization (cBO)~\citep{Metzen2015} which extends the classic BO framework from \sec{sec:bg:BO}.
	Contextual Bayesian Optimization learns a joint model $\tilde{\objfuncNo}: \{\parameters,\context\} \rightarrow \objfunc{\parameters}$, but now, at every iteration the acquisition function is optimized with a constrained optimization where the context $\context$ is provided by the environment. 
	However, because the model jointly model the context-parameter space, experience learned in one context can be generalized to similar contexts. 
	By utilizing cBO, we will show in \sec{sec:results} that our microrobot can learn to walk (and generalize) to different environmental contexts such as walking uphill and curving.

%% file: 5_approach.tex
\begin{figure}[t]
	\centering
	\includegraphics[height=2.4cm]{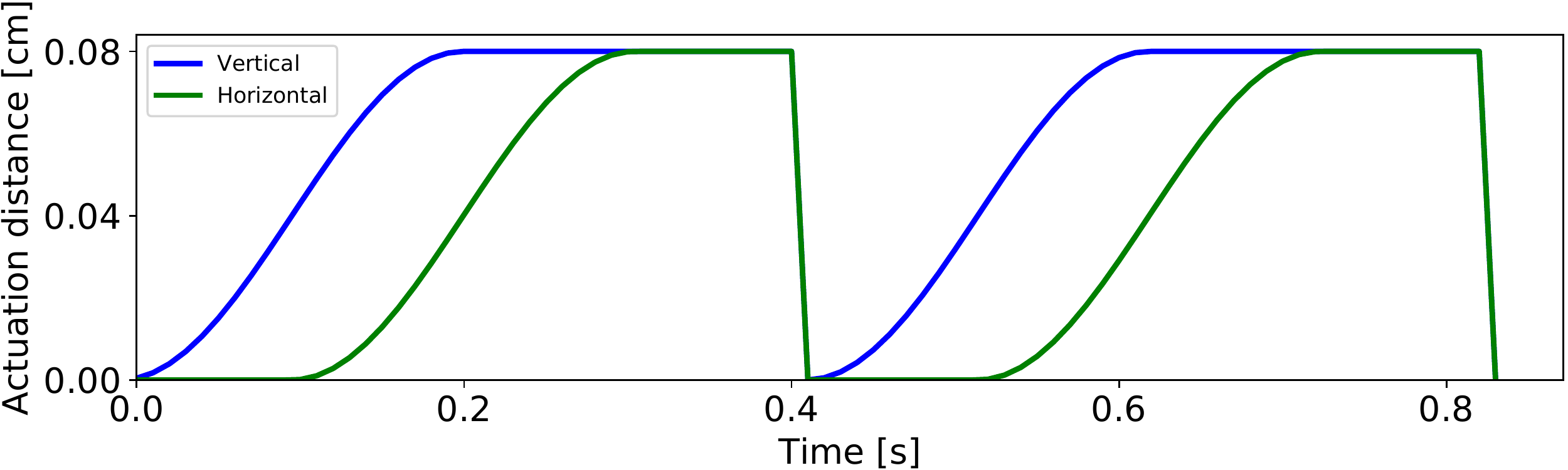}
	\caption{Output of one vertical-horizontal oscillator pair in the CPG network, which corresponds to one leg on the robot. 
	The retraction phase of both motors occurs concurrently and rapidly in order to simulate the physical constraints on the actual physical microrobot.}
	\label{fig:cpg}
\end{figure}
\begin{figure}[t]
	\centering
	\includegraphics[width=0.98\linewidth]{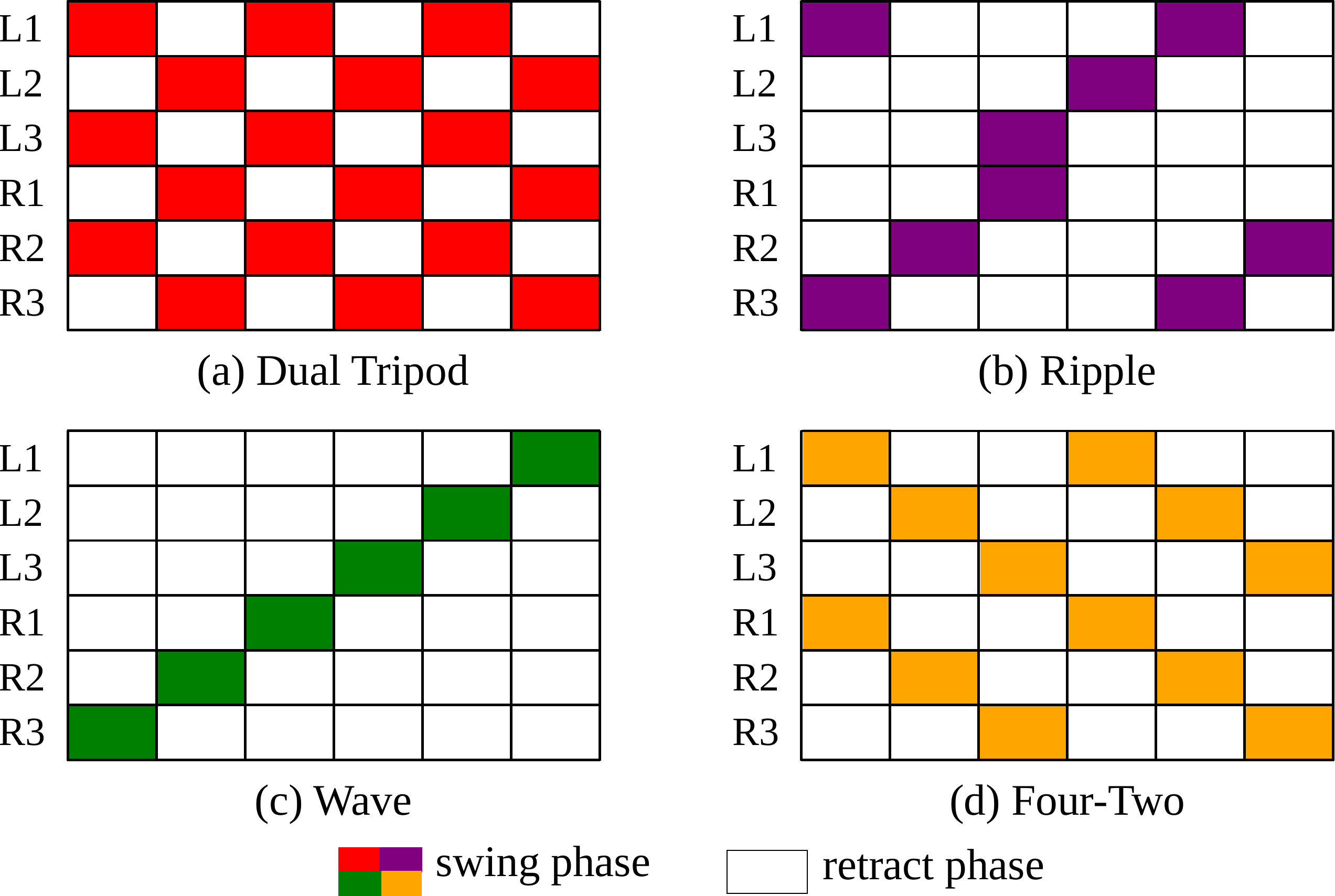}
	\caption{Contact/swing patterns for different gaits.}
	\label{fig:gaits}
	\vspace{-8pt}
\end{figure}

We now present our novel approach to learn motor primitives for path planning.
This approach relies on the possibility of re-using the evaluations collected using cBO to convert the task into a multi-objective optimization problem.
We specifically consider a cBO task where we want to optimize the parameters $\parameters$ to reach different target positions $\context=\left[ \Delta x_\text{des}, \Delta y_\text{des} \right]$ (this setting is evaluated in \sec{sec:results:context2}).
The objective function in this case can be defined as the Euclidean distance 
\begin{align}
	\objfuncNo = \sqrt{\left(\Delta x_\text{des} -\Delta x_\text{obs}\right)^{2} + \left(\Delta y_\text{des} -\Delta y_\text{obs}\right)^{2}}\,,
\end{align}
where $\Delta x_\text{obs}, \Delta y_\text{obs}$ are the actual positions measured after evaluating a set of parameters.
The cBO model would map $\tilde{\objfuncNo}: \left[\parameters, \Delta x_\text{des}, \Delta y_\text{des}\right] \rightarrow \objfunc{\parameters}$. 
However, in order to compute $f$ it would need to measure $\Delta x_\text{obs}, \Delta y_\text{obs}$, effectively generating data of the form
\begin{align}
	\left[\parameters, \Delta x_\text{des}, \Delta y_\text{des}\right] \rightarrow \left[\Delta x_\text{obs}, \Delta y_\text{obs}, \objfunc{\parameters}\right]
\end{align}
We can now re-use the data generated from this contextual optimization to learn a motor primitive model in the form $g: \parameters \rightarrow \left[\Delta x_\text{obs}, \Delta y_\text{obs}\right]$.
The purpose of this learned model~$g$ is now to provide an estimate of the final displacement obtained for a set of parameters independently from the optimization process that generated it.
Once such a model is learned, we can use it to compute parameters that lead to the desired displacement $\Delta x^*_\text{obs}, \Delta y^*_\text{obs}$ by optimizing the parameters w.r.t. the output of the model
\begin{align}
	\parameters^* = \maximize_{\parameters}\, z(g(\parameters))\,,
\end{align}
where $z$ is a scalarization function of our choice (e.g., the Euclidean distance).
This is equivalent to learning a continuous function that generates motor primitives from the desired displacement.
It should be noted that this optimization is performed on the model $g$ and therefore does not require any physical interaction with the robot.
Moreover, we can optimize the parameters over a series of multiple displacements to obtain a path planning optimization. 
In \sec{sec:results:planning}, when performing path planning using the learned motor primitives we will employ a simple shooting method optimization which randomly samples multiple candidate parameters and selects the best outcome.

%% file: 6_results.tex
\begin{figure}[t]
  \centering
  \begin{subfigure}{0.49\linewidth}
	  \includegraphics[width=0.98\linewidth]{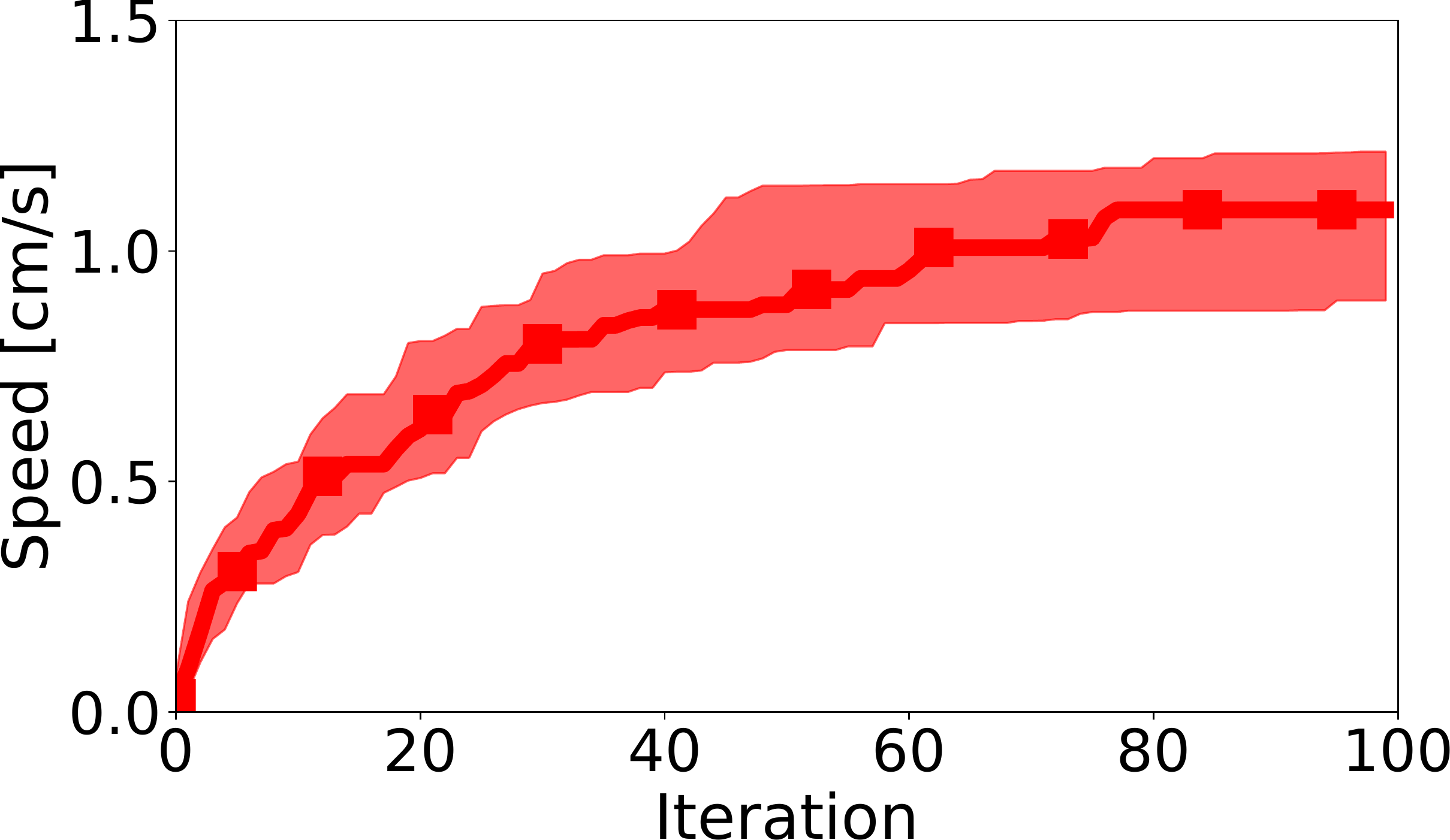}
	  \caption{Dual Tripod}
	  \label{fig:soo:1}
  \end{subfigure}
  \hfill  
  \begin{subfigure}{0.49\linewidth}
	  \includegraphics[width=0.98\linewidth]{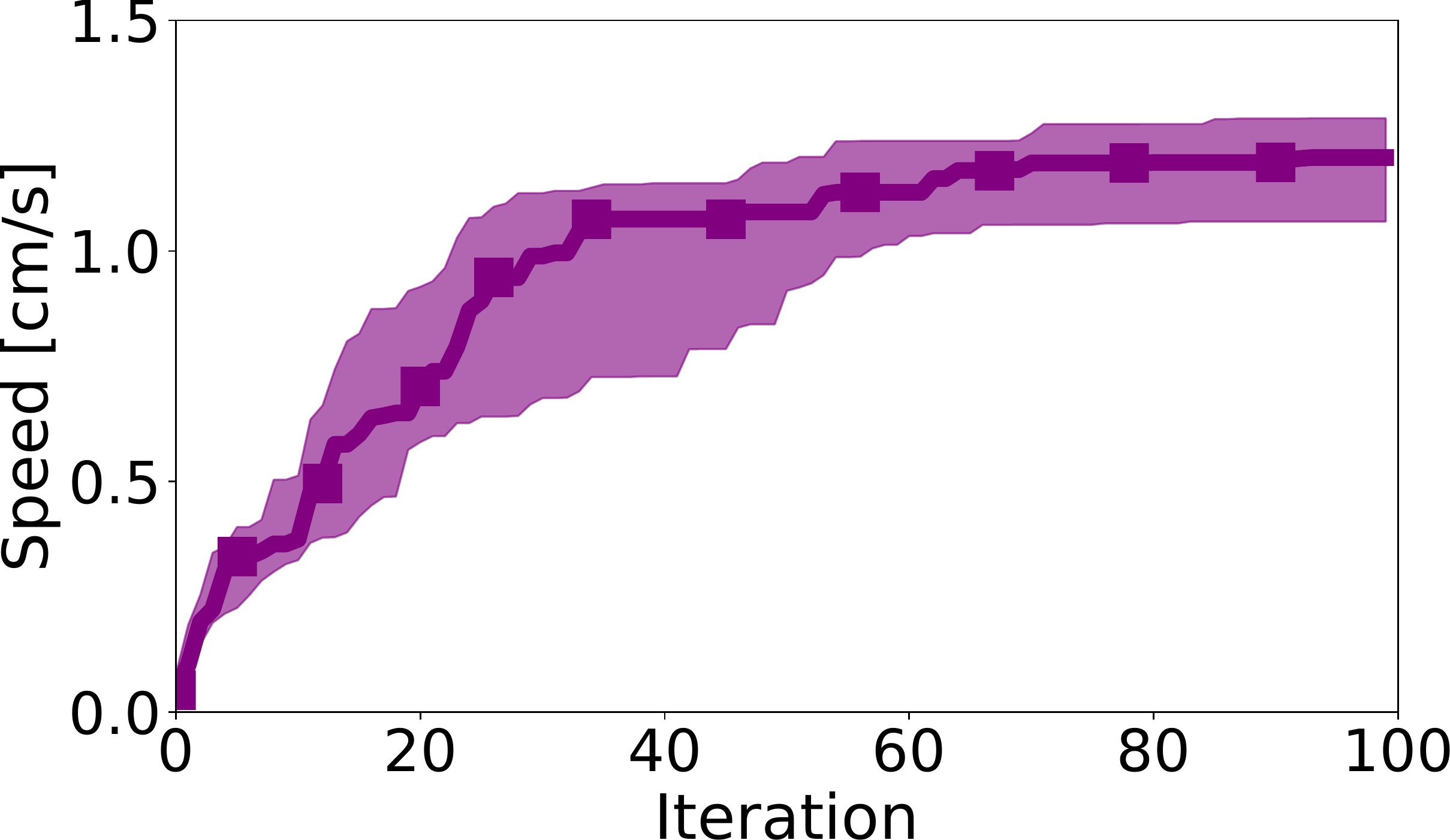}
	  \caption{Ripple}
	  \label{fig:soo:2}
  \end{subfigure}
  \\
  \begin{subfigure}{0.49\linewidth}
	  \includegraphics[width=0.98\linewidth]{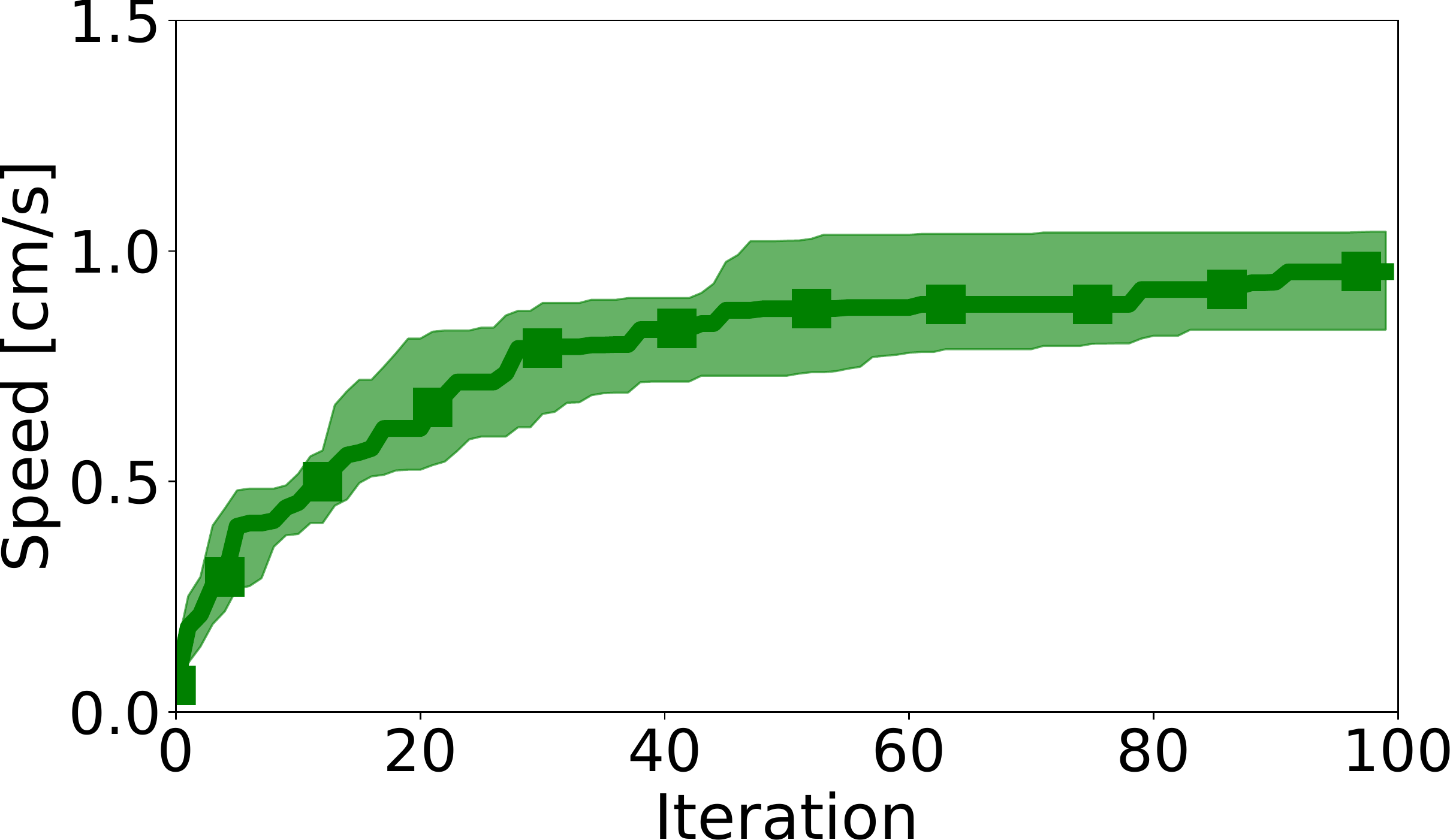}
	  \caption{Wave}
	  \label{fig:soo:3}
  \end{subfigure}
  \hfill
  \begin{subfigure}{0.49\linewidth}
	  \includegraphics[width=0.98\linewidth]{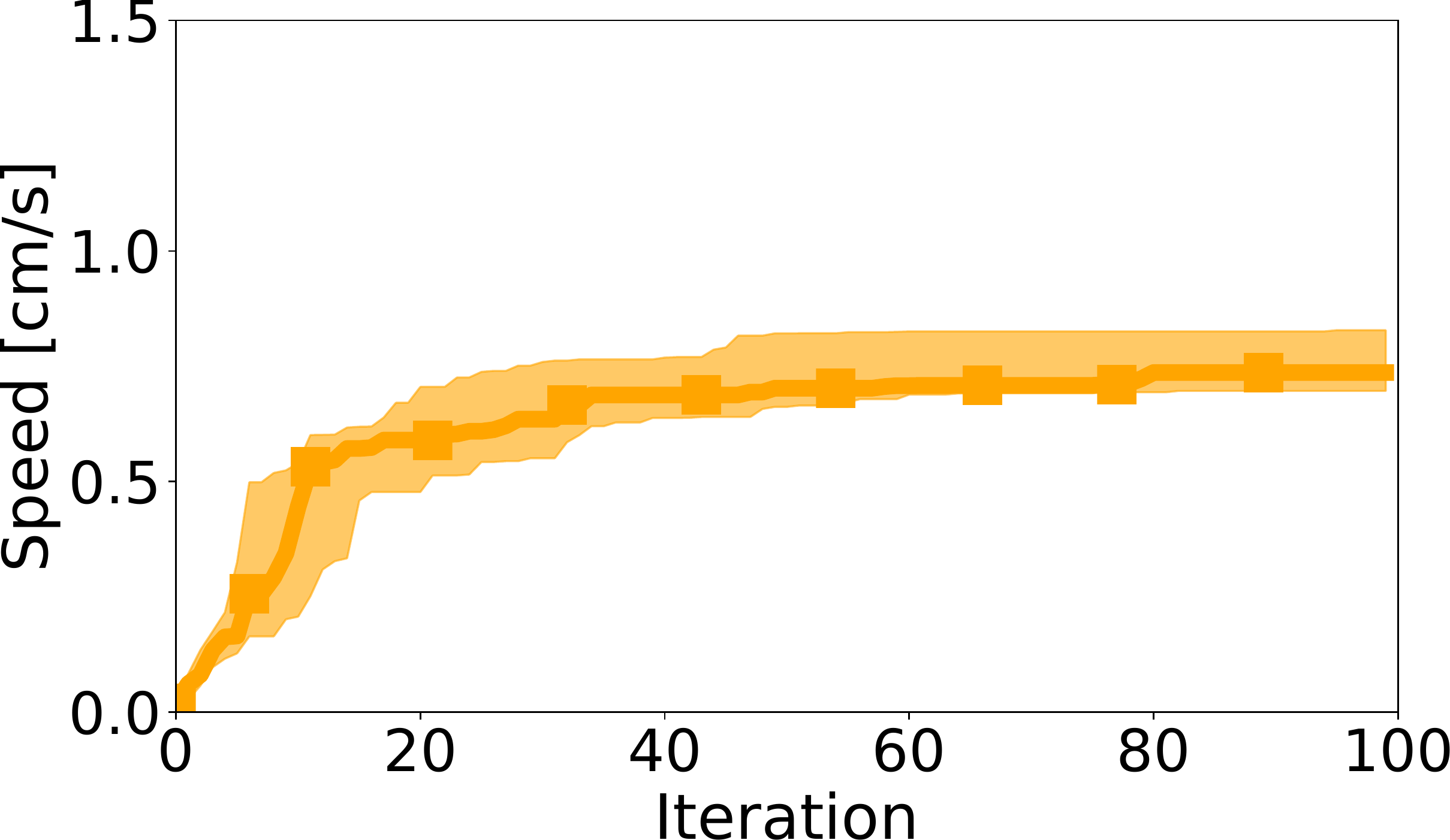}
	  \caption{Four-Two}
	  \label{fig:soo:4}
  \end{subfigure}
  \caption{Learning curve for the four gaits (median and 65th percentile). We can see how, for all the gaits, BO learns to walk from scratch within 50 iterations. After the optimization, Dual Tripod and Ripple are the fastest gaits at $\sim \SI[per-mode=symbol]{1.1}{\centi\meter\per\second}$ and $\sim \SI[per-mode=symbol]{1.2}{\centi\meter\per\second}$ respectively.}
  \label{fig:soo}
\end{figure}
In this section, we discuss our controller implementation as well as the performance of our simulated microrobot on various locomotion tasks.
The code used for performing the simulation and videos of the various locomotion tasks are available online at \url{https://sites.google.com/view/learning-locomotion-primitives}.

\subsection{Controller Implementation}
	We built our controller following the setup described in \sec{sec:bg:cpg}, using a network of 12 coupled phase oscillators (one per motor).
	In order to translate the output of each of the oscillators into motor actuation, we calculate the oscillator outputs for each vertical-horizontal motor pair using the piecewise function
	\begin{align}
		    \begin{cases}
			    x_{i} + r_{i}cos(\phi_{i}), x_{j} + r_{j}cos(\phi_{j}) &\text{if }\phi_{i}>\pi,\phi_{j}>\pi\,,\\
			    x_{i} + r_{i}, x_{j} + r_{j}cos(\phi_{j}) &\text{if }\phi_{i}\leq\pi,\phi_{j}>\pi\,,\\
			    x_{i} + r_{i}, x_{j} + r_{j} &\text{if }\phi_{i}\leq\pi,\phi_{j}\leq\pi\,,\\
			    x_{i} + r_{i}cos(\phi_{i}), x_{j} - r_{j} &\text{if }\phi_{i}\leq\pi,\phi_{j}>\pi\,,
		    \end{cases}
	\end{align}
	where the $i$th oscillator outputs to its respective vertical motor and the $j$th oscillator outputs to its respective horizontal motor. 
	This allows us to discard the parts of the oscillator output that are not consistent with the physical constraints of the physical robot, since the actual leg actuators cannot partially retract (see \fig{fig:cpg}).
	We choose to mutually couple all six of the vertical oscillators (with a coupling weight of 4 to ensure quick convergence on stable limit cycles).
	We refer the reader to \cite{Crespi_2007} for a more comprehensive discussion of oscillator coupling in CPGs.
	Each of the horizontal oscillators are also coupled with their respective vertical oscillator in order to encapsulate the dynamics of each leg.
	We chose to implement four different gaits with the CPG -- tripod, ripple, wave, and four-two (see \fig{fig:gaits}). 
	For a more detailed description of these gaits we refer the reader to~\cite{Campos2010}.
	We use the same frequency and phase difference for the whole network in order to reduce the number of parameters and speed up the rate of convergence.
	We use two separate parameters for amplitude, each controlling the left and right set of legs respectively.
	This choice of parameters allows us to control the turning of the robot which is necessary for path planning and corrections for not walking straight.

\subsection{Learning to Walk Straight}
\label{sec:results:soo}

	We optimized the four gaits considered (i.e., dual tripod, ripple, wave, and four-two) using as our objective function the walking speed of the robot (measured as the distance traveled after $\SI{1}{\second}$).
	Since some gaits result in curved motions, we also penalized the speed objective with a term proportional to the drift from the axis of locomotion.
	The optimization used the 4 parameters outlined in \sec{sec:bg:cpg} and was repeated 50 times for each of the gaits. 
	In \fig{fig:soo}, we show the median and 65th percentiles of the best solution obtained so far in the trials.
	The results show that the optimizer was able to learn to walk from scratch within 50 iterations.
	Moreover, it can be noted that the optimized tripod and ripple are the fastest gaits at $\sim \SI[per-mode=symbol]{1.1}{\centi\meter\per\second}$ and $\sim \SI[per-mode=symbol]{1.2}{\centi\meter\per\second}$ respectively.
	
\subsection{Multi-objective Gait Optimization}
\label{sec:results:moo}
	\begin{figure}[t]
	  \centering
	  \begin{subfigure}{0.49\linewidth}
		  \includegraphics[width=0.99\linewidth]{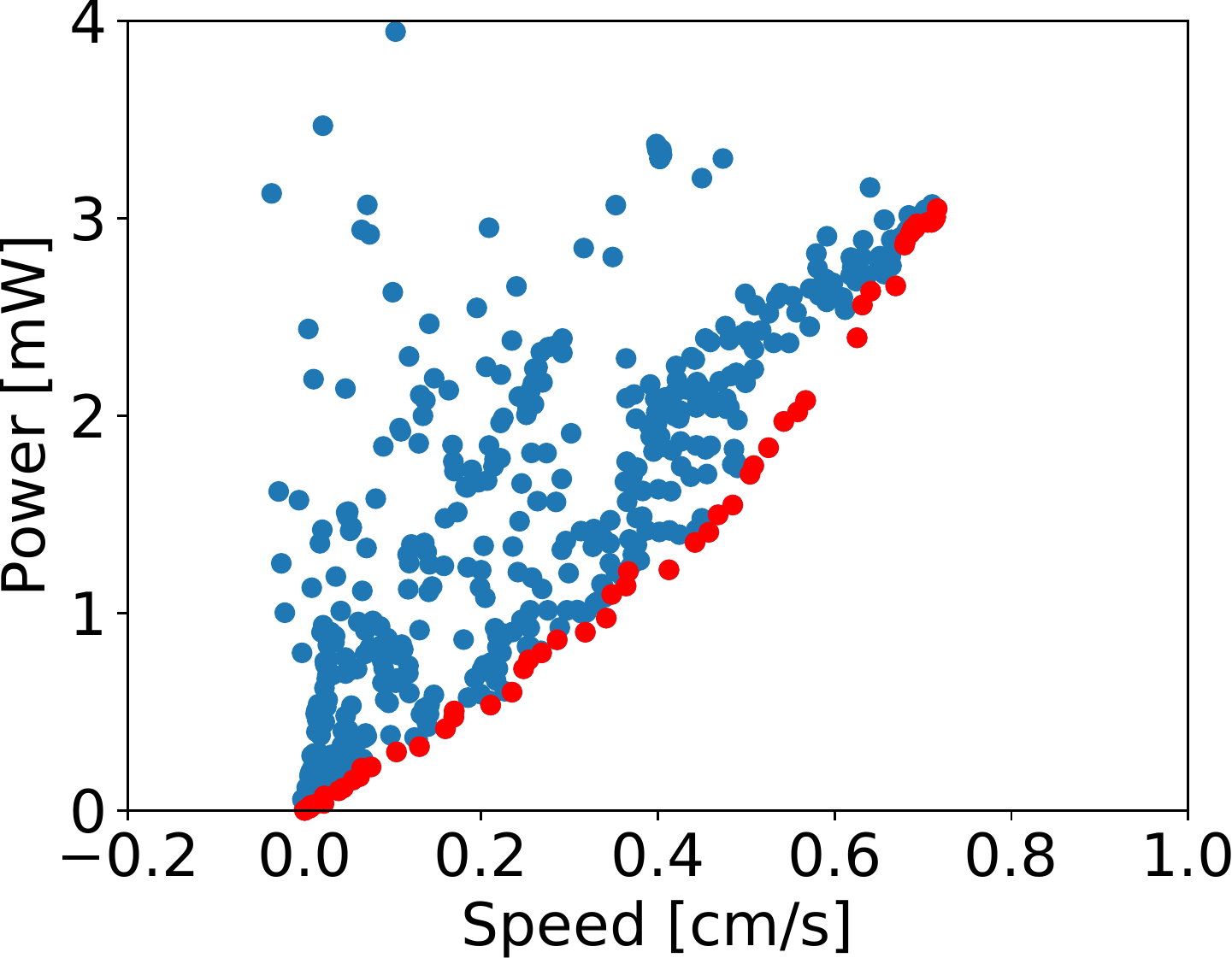}
		  \caption{Dual Tripod}
		  \label{fig:moo:1}
	  \end{subfigure}
	  \hfill  
	  \begin{subfigure}{0.49\linewidth}
		  \includegraphics[width=0.99\linewidth]{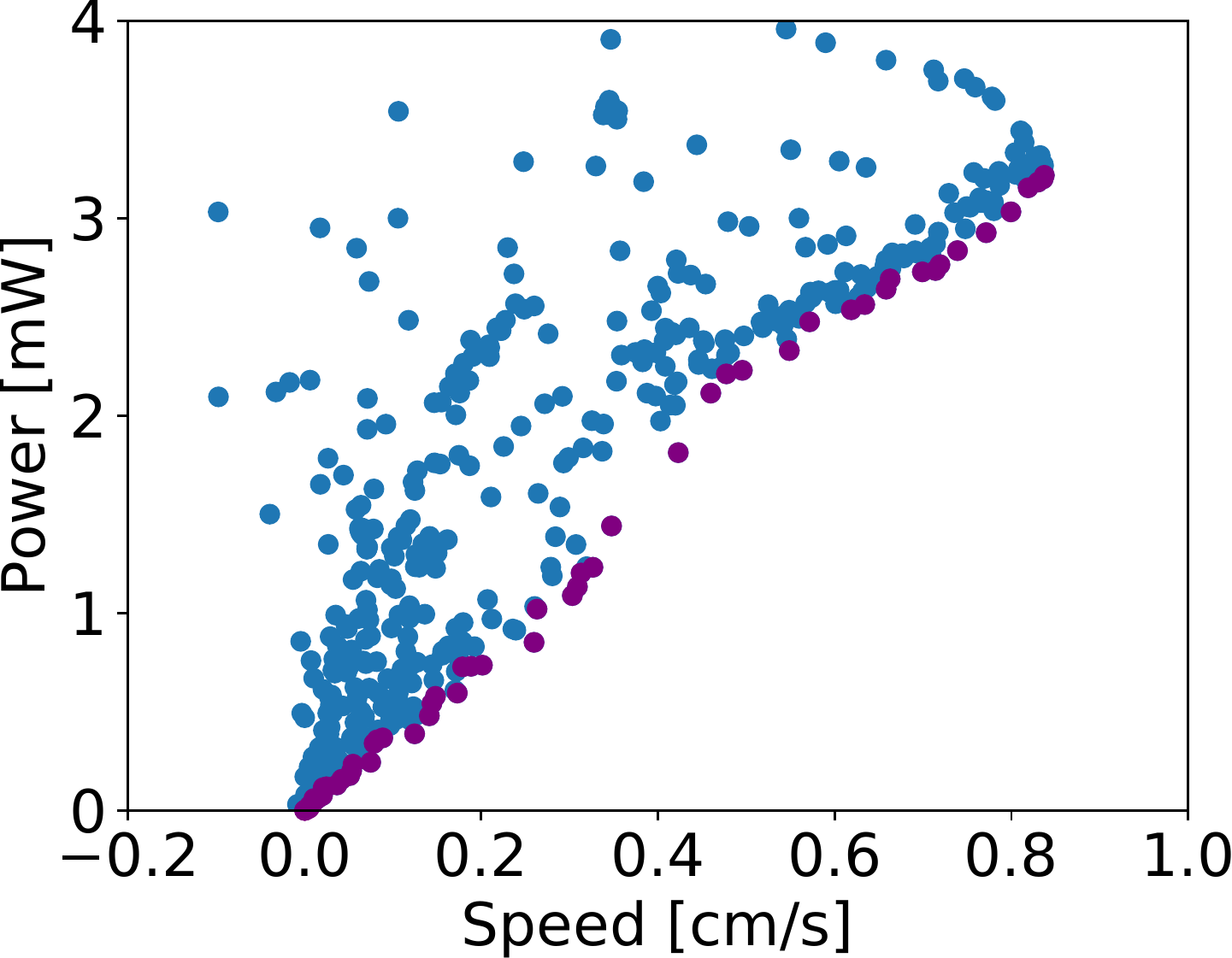}
		  \caption{Ripple}
		  \label{fig:moo:2}
	  \end{subfigure}
	  \\
	  \begin{subfigure}{0.49\linewidth}
		  \includegraphics[width=0.99\linewidth]{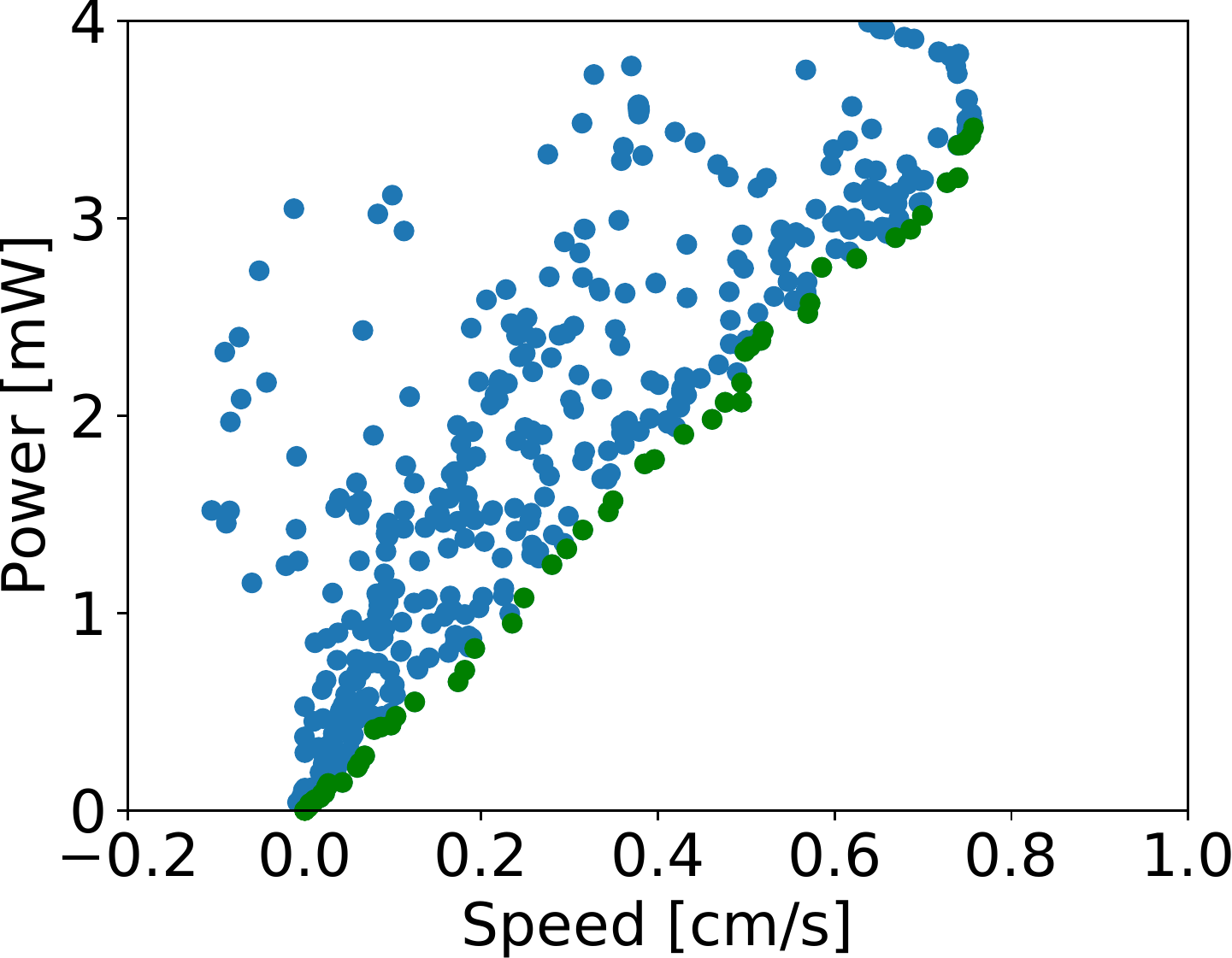}
		  \caption{Wave}
		  \label{fig:moo:3}
	  \end{subfigure}
	  \hfill
	  \begin{subfigure}{0.49\linewidth}
		  \includegraphics[width=0.99\linewidth]{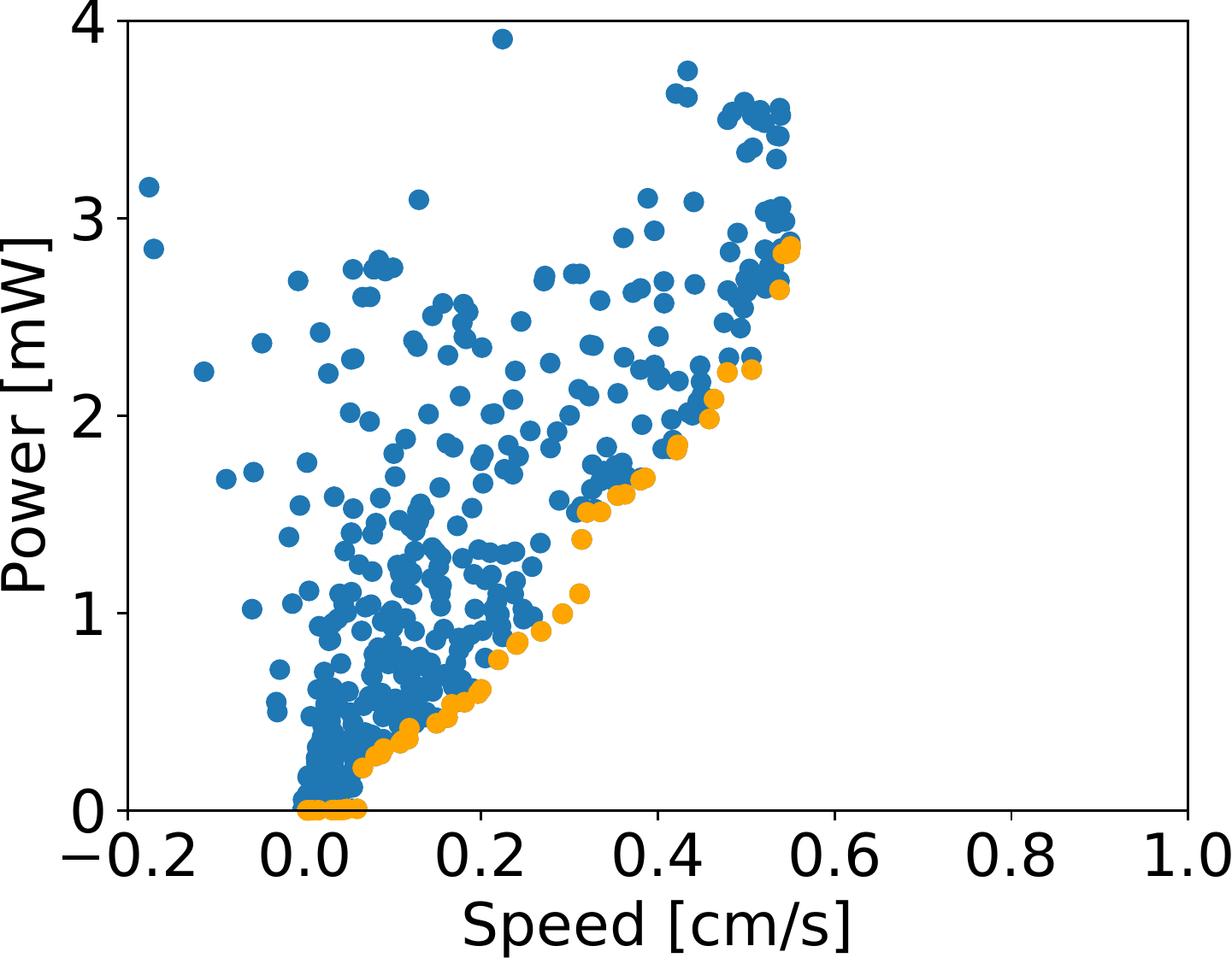}
		  \caption{Four-Two}
		  \label{fig:moo:4}
	  \end{subfigure}
	  \caption{Performance measured for the four gaits, and the corresponding PFs. ParEGO is able to quickly explore the PF for each of our four gaits.}
	  \label{fig:moo}
	\end{figure}

	In the previous simulation we only considered walking speed as our objective. 
	However, for practical gait design, energy efficiency is another objective of great interest, particularly when it comes to designing gaits for a microrobot with real energy restrictions. 
	For this reason, we now consider a multi-objective optimization setting and compare the different gaits w.r.t. both walking speed, and energy consumption.
	The energy consumption of the robot was computed by measuring the forces exerted by each of the 12 motors along the axis of actuation and calculating the power used to actuate the motors. 
	Since the retraction of the legs is spring powered, the energy input in the cycle is only during motor extension.
	Hence, we only consider the cost of extending the legs.
	With the mass of the robot and the time of each trial being held constant, we quantify the energy efficiency of a gait and estimate the cost of transport.

	\begin{wrapfigure}{r}{0.52\linewidth} 
	\vspace{-12pt}
	  \centering
	  \includegraphics[width=\linewidth]{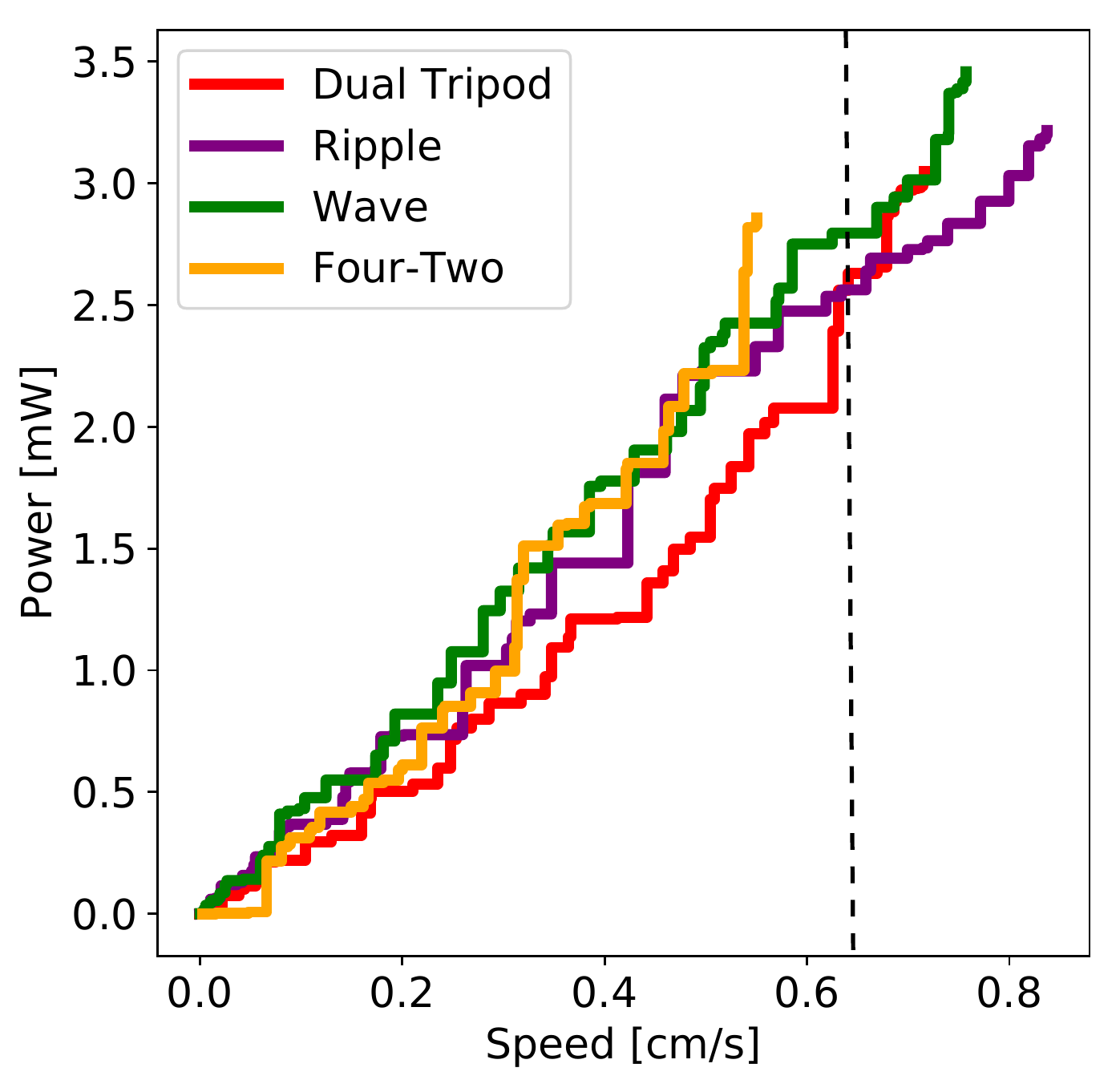}
	  \caption{Comparison of the PFs obtained for the different gaits.}
	  \label{fig:moo:all}
	  \vspace{-10pt}
	\end{wrapfigure}
	We optimized the four gaits again with the same 4 parameters as the previous optimization, but this time using multi-objective Bayesian optimization with a budget of 50 iterations.
	\begin{figure}[t]
	  \centering
	  \includegraphics[width=0.95\linewidth]{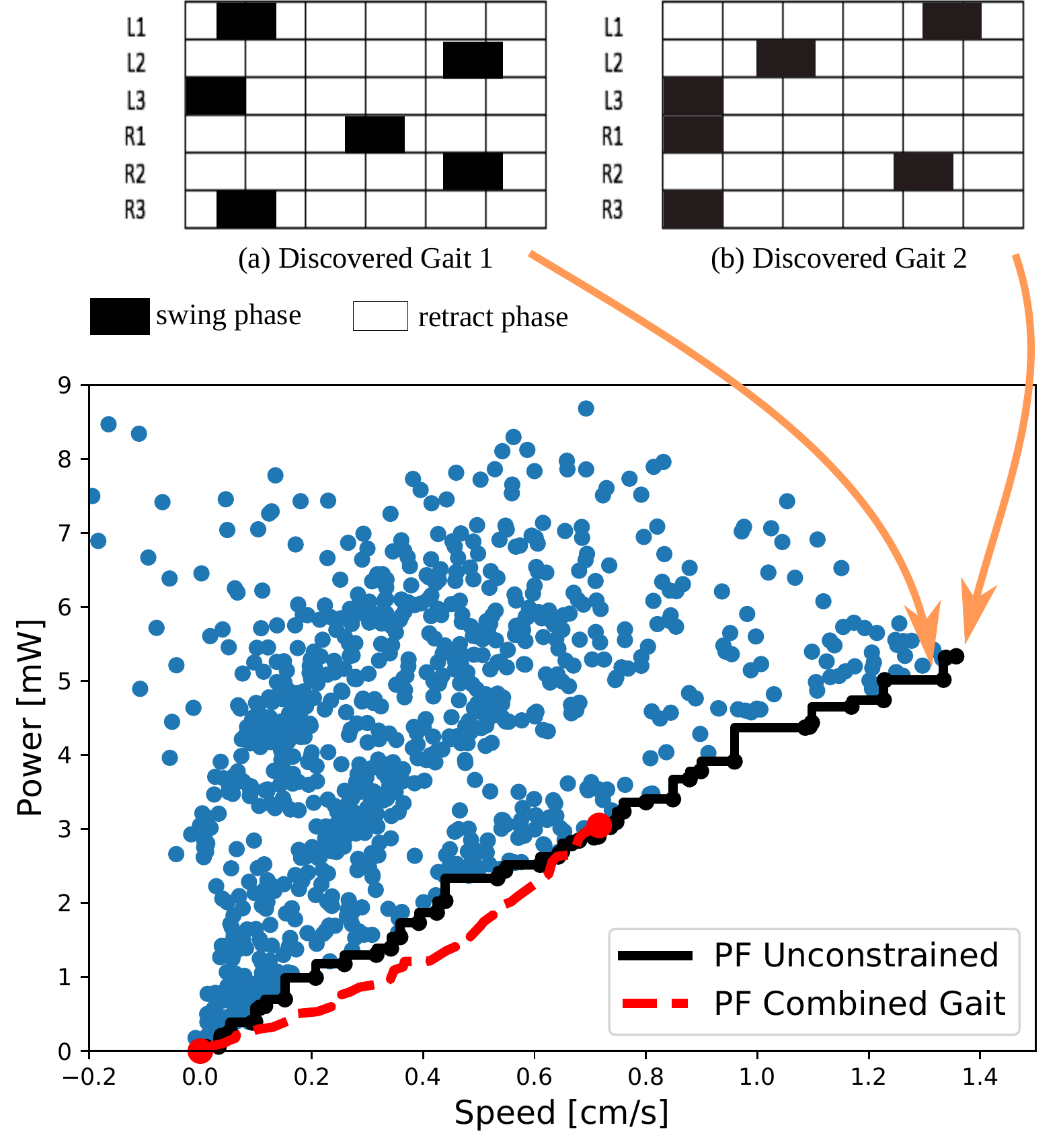}
	  \caption{PF of the unrestrained gait optimization versus the best performance of the four nature-inspired gaits. The faster solutions outperform the fastest nature-inspired gaits, albeit with more energy expenditure. However, the inability of the optimizer to match the performance of the gaits at lower speeds within 1250 trials shows that the gait parametrization can help limit the search space to find better solutions easier. \textit{(top)} Pattern for two of the discovered gaits.}
	  \label{fig:moo:new}
	\end{figure}
	In \fig{fig:moo} we can see the performance measured and Pareto fronts obtained for the different gaits.
	To better compare the PF from the different gaits, we also visualized just  the PFs together in \fig{fig:moo:all}. 
	From these results, we can see how the tripod gait dominates the other gaits for speed $<\SI[per-mode=symbol]{0.6}{\centi\meter\per\second}$, while Ripple dominates when the speed is $>\SI[per-mode=symbol]{0.6}{\centi\meter\per\second}$, hence giving a clear indication of which gait is preferable under different circumstances.

\subsection{Discovering New Gaits with Multi-objective Optimization}

	In addition to optimizing the four nature-inspired gaits, we also tested multi-objective optimization on the walker without constraining to using predefined gaits.
	To parametrize the oscillator couplings, we thus discretized each gait into intervals of constant length.
	Within each of these intervals, we assume that each leg steps exactly once, keeping each of the oscillators in the CPG in phase with each other.
	This allows us to parametrize gaits by assigning each leg a point during each interval where it begins stepping.
	While this parametrization excludes certain gaits that cannot be expressed in this form, we leave the study of more sophisticated gait parameterizations for gait discovery to future works.
	
	The resulting multi-objective optimization task had 8 parameters (frequency, phase difference between horizontal and vertical motors, and the six gait coupling parameters).
	Due to the higher parameter dimensionality, and because this training was not intended for on-line training, we ran the optimization for 250 iterations in order to allow a more comprehensive exploration of the optimization space.
	We also repeated the optimization five times for a total of 1250 trials.
	In \fig{fig:moo:new} we can see the Pareto front for the resulting gaits.
	We found that the fastest discovered gaits were actually able to outperform the four nature-inspired gaits implemented by a substantial margin.
	Even while penalizing curved paths, the fastest discovered gait outperformed Ripple (the fastest nature-inspired gait we found) by almost $50\%$.
	However, for low-speed gaits, the nature inspired gaits out-perform the gaits produced by the unconstrained optimization, indicating the optimization did not yet fully converged to the optimal PF.
	
\subsection{Learning to Walk on Inclined Surfaces}
\label{sec:results:context1}
	\begin{figure}[t]
	  \centering
	  \includegraphics[width=0.96\linewidth]{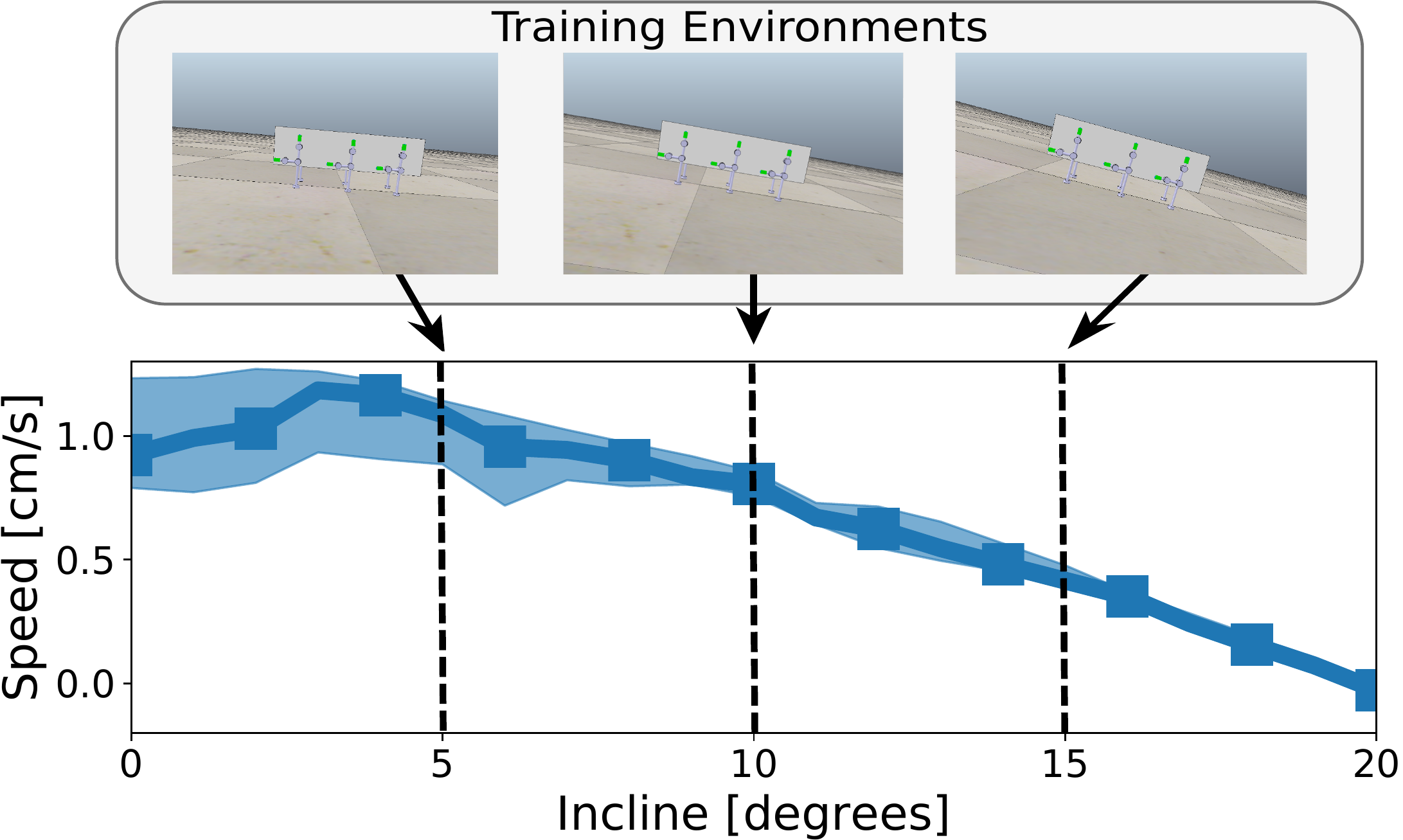}
	  \caption{Performance of the contextual policy (median and 65th percentile) for a wide range of inclines. The policy was trained only at 5, 10 and 15 degrees, but it was capable of generalizing smoothly to unseen inclinations. 
	  }
	  \label{fig:incline}
	\end{figure}
    
	\begin{figure}[t]
	  \centering
	  	\begin{subfigure}{0.49\linewidth}
      \centering
	  \includegraphics[width=\columnwidth]{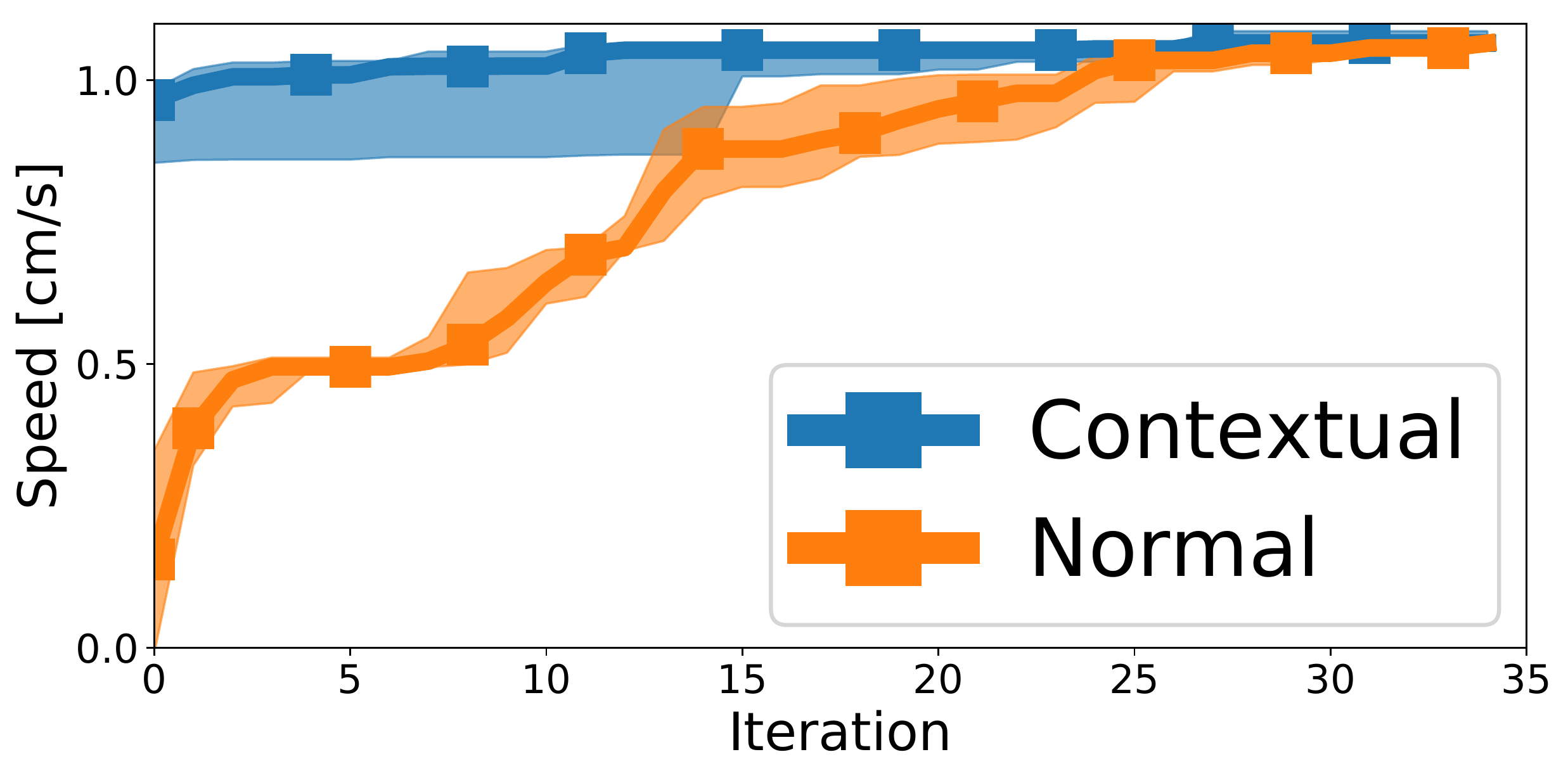}
	  \caption{Inclined surface.}
	  \label{fig:contextual:1}
	\end{subfigure}
\hfill
	\begin{subfigure}{0.49\linewidth}
	  \centering
	  \includegraphics[width=\linewidth]{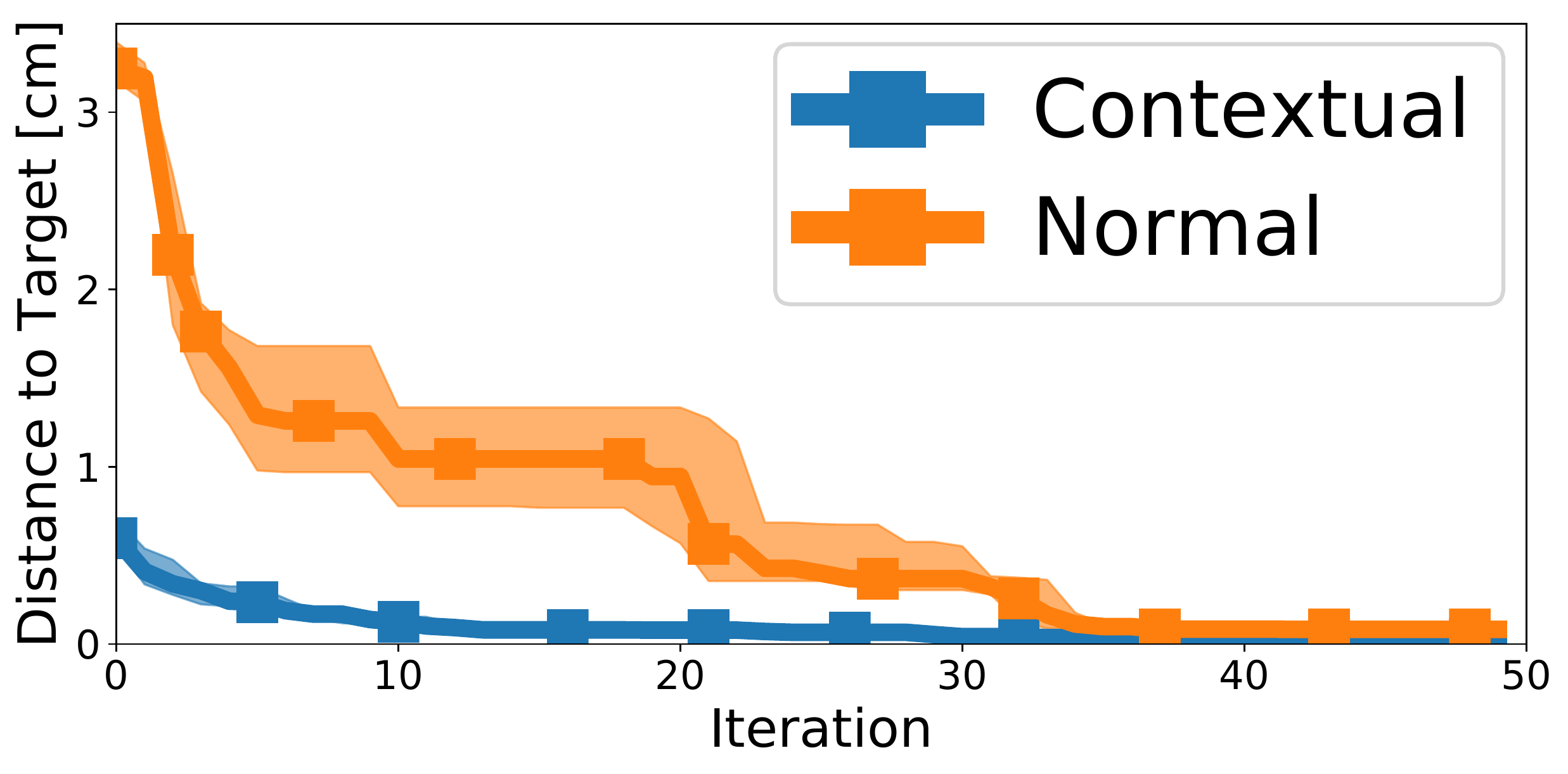}
	  \caption{Curved trajectory.}
	  \label{fig:contextual:2}
	  	\end{subfigure}
\caption{Comparison between the optimization performance of a contextual optimizer and a normal optimizer for two different tasks: (a) walking on inclines (b) walking curved trajectories. In both cases, the contextual optimizer can leverage prior simulations to obtain high-performing gaits in fewer simulations.}
	   \vspace{-10pt}
       \end{figure}
       
	We now consider the case of contextual optimization and specifically the task of gait optimization for slopes with different inclinations.
	We framed learning to walk on inclined terrain as a contextual policy search, where the angle of the inclination is the context. 
	In this simulation, we decided to use Dual Tripod for our gait with mostly the same open parameters as the previous simulations.
    We used a single parameter to represent the amplitude for the entire network in order to keep the number of parameters low with the addition of a contextual variable, leaving us with 3 parameters and 1 contextual parameter.
	To respect real world constraints, where testing randomly sampled incline angles over a continuous interval can be excessively time-consuming, we chose at training time to perform simulations only from a small number of inclines: 5, 10, and 15 degrees.
	
	After optimizing the gaits for these three inclines over 50 iterations, we studied how the contextual optimizer is able to generalize across the context space by testing the performance of the contextual policy for a wide range of inclines.
	In \fig{fig:incline} we can see that the policy performs well on intermediary inclines and seems to smoothly interpolate between the training inclines as is desirable.
	The gradual decrease in performance as the inclines get steeper can be attributed to the increasing physical difficulty for climbing up steeper inclines.
	We also compared cBO against using standard BO to train the robot for an untested incline. 
	As shown in \fig{fig:contextual:1}, the contextual optimization was able to converge on optimal performance significantly faster than standard BO.
	This result demonstrate the ability of cBO to efficiently use data accumulated in previous contexts to quickly reach optimize gaits in new unseen contexts.

\subsection{Learning to Curve}
\label{sec:results:context2}
	
	Another useful task that can be framed as contextual optimization is learning motor primitives to walk curved trajectories for use in path planning.
	We used the same parameters as in \sec{sec:results:soo} and the contextual parameters in this case were the target displacements along both the x and y axes from the point of origin.
	In order to train particular trajectories, we selected five evenly spaced target points along the front quadrant of the field of vision.
	Since the primary objective was to reach the desired destination, we chose to use the distance of the final position to the target position as our sole objective function.
	We found that over 10 repetitions, the walker was able to accurately move and turn towards all of the target points within 250 iterations.
	In \fig{fig:contextual:2}, we compared the performance of cBO against standard BO on a previously unseen target position $(4\cos{\pi / 16}, 4\sin{\pi / 16})$.
	We found that, as in the case of inclinations, the contextual policy was able to learn the optimal parameters for a novel trajectory within very few iterations.

\subsection{Learning Motor Primitives for Path Planning}
\label{sec:results:planning}
	\begin{figure}[t]
	  \centering
	  \includegraphics[width=\linewidth]{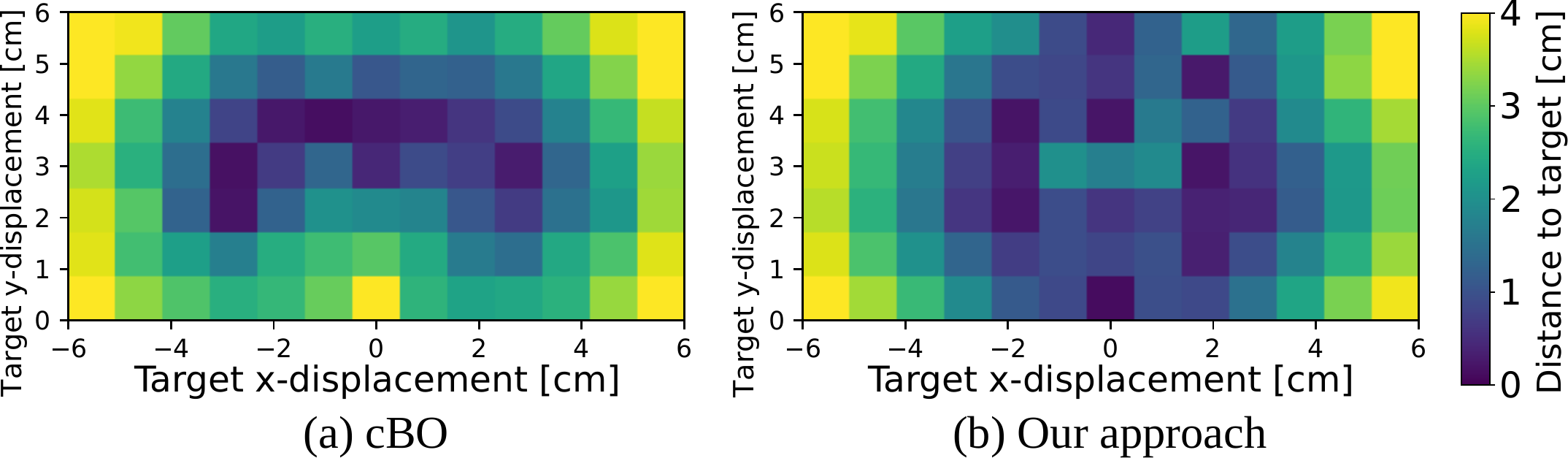}
	  \caption{Comparison of the performances of cBO and our approach for learning motor primitives (using the same data). 
	  With the robot having an initial position of $(0,0)$, we evaluated the error between the desired position (indicated by the element of the grid) and the reached position.
	  Darker color indicates better target accuracy.
	  While cBO accurately learned trajectories near the training targets, it did not generalize well to unseen targets. 
	  In contrast, our approach had a more comprehensive coverage as it could leverage better information about the environment to improve generalization.}
	  \label{fig:pathing1}
	\end{figure}
	\begin{wrapfigure}{r}{0.50\linewidth} 
	\vspace{-10pt}
	  \centering
	  \includegraphics[width=\linewidth]{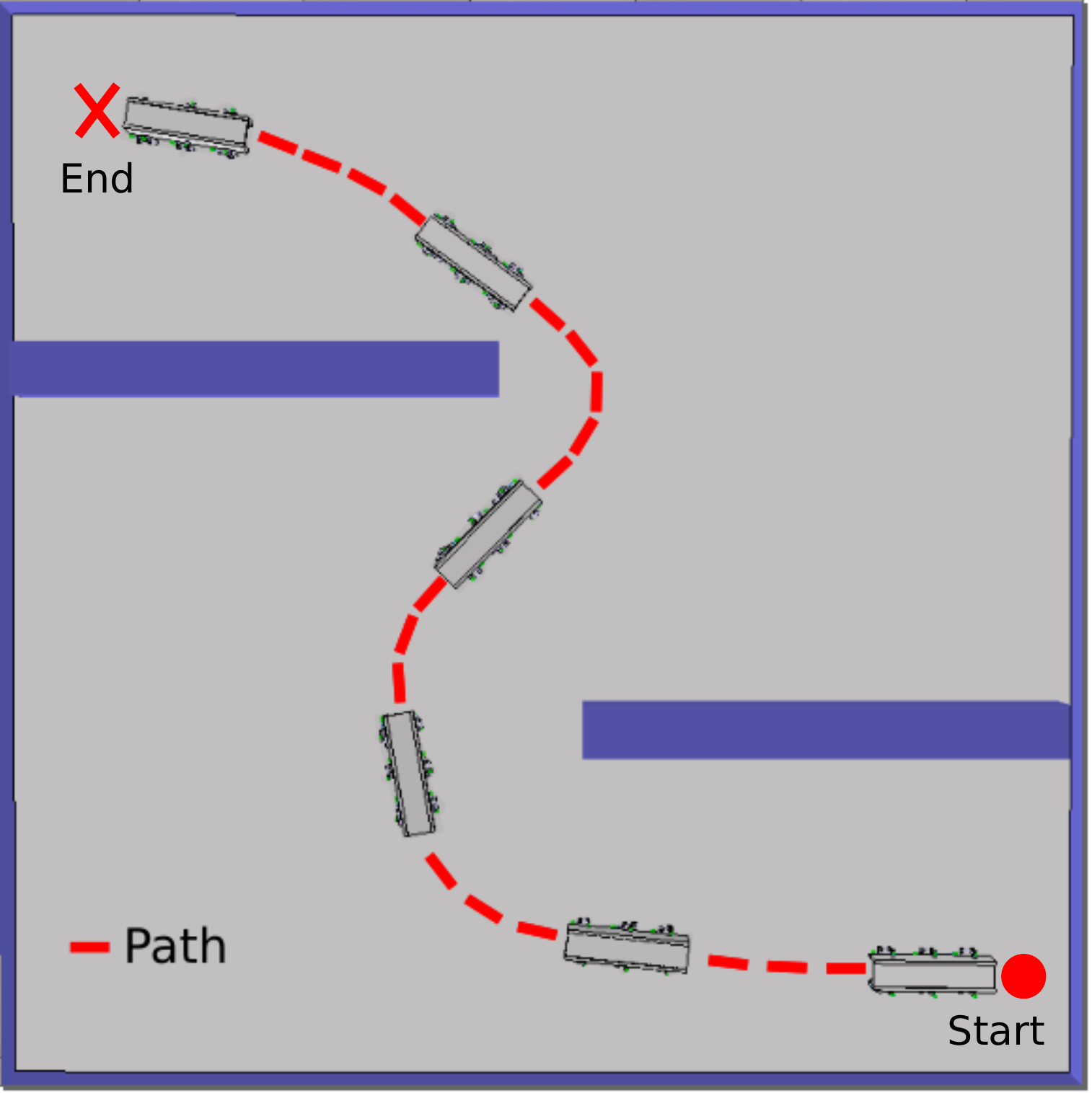}
	  \caption{Path constructed using the locomotion primitives learned with our approach.}
	  \label{fig:pathing2}
	   \vspace{-8pt}
	\end{wrapfigure}
	In the previous simulation we learned motor primitives capable of walking curved trajectories.
    While the model handled trajectories near and between the targets quite well, the performance on trajectories well within the physical capabilities of the robot but not in proximity to the targets left much to be desired, as shown in \fig{fig:pathing1}.
	We now demonstrate how our approach presented in \sec{sec:approach} can be used to significantly improve the movement accuracy (compared to cBO using the same data), as well as how such motor primitives can be used to perform path planning.
    First, we reused the data from the previous simulation in order to reformulate the task as a multi-objective optimization as described in \sec{sec:approach}.
    Then, we used our trained model to sample 10,000 trajectories by randomly sampling from the parameter space.
    Out of all these trajectories, we selected the one with the smallest expected error subject to not walking through the wall.
    Evaluating the resulting sequence of motor primitives on the real system (\ie, the simulator) demonstrated that the expected trajectory was capable of navigating the maze, as shown in \fig{fig:pathing2}.

%% file: 7_conclusion.tex
Designing controllers for locomotion is a daunting task.
In this paper, we demonstrated on a simulated microrobot that this process can be significantly automated. 
Our main contributions are two-fold: 
1) we introduced a coherent curriculum of increasing challenging tasks, which we use to evaluate the CPG controller of our microrobot using Bayesian optimization.
2) we presented a new approach that enables walking robots to efficiently learn motor primitives from scratch.
By using the data collected from contextual optimization we reformulate the problem into a multi-objective optimization task, and learn a model that can map any set of parameters to a predicted trajectory. 
This model can subsequently be used for path planning.
Our experimental simulation results demonstrate that using this approach a microrobot can successfully learn accurate locomotion primitives within 250 trials, and subsequently use them to navigate through a maze, without any prior knowledge about the environment or its own dynamics. 

The gaits obtained on the simulated microrobot might not yield good results when applied to the real microrobot, due to the low-fidelity of the simulator used. 
However, the methodology used to obtain them is realistically applicable to real microrobots, and is uniquely able to address concerns that exist on the sub-centimeter scale (\eg, lack of a precise physics simulator and budgeting of physical experiments). 
In future work, we plan to evaluate our approach and findings on the physical hexapod microrobot.